\documentclass[11pt]{article}

\usepackage[preprint]{acl}

\usepackage{times}
\usepackage{latexsym}
\usepackage{amsmath}

\usepackage{booktabs}
\usepackage{makecell}

\usepackage{csquotes}

\usepackage{tikz}
\usetikzlibrary{shapes.geometric,shapes.symbols,arrows.meta,positioning,fit,shadows,backgrounds}
\usetikzlibrary{calc} 
\usetikzlibrary{patterns}
\newcommand{\rewritefigurescale}{0.70}

\usepackage{pgfplots}
\pgfplotsset{compat=newest}

\usepackage[T1]{fontenc}

\usepackage[utf8]{inputenc}

\usepackage{microtype}

\usepackage{inconsolata}

\usepackage{graphicx}

\usepackage{latexml}
\newcommand{\htmlfig}[2]{\includegraphics[width=\linewidth,alt={#2}]{latex/figures/svg/#1.svg}}
\iflatexml
  \makeatletter\newif\ifacl@anonymize\makeatother
  \providecommand{\Needspace}[1]{}
  \lxRequireResource{htmlfix.css}
\fi
\DeclareRobustCommand{\notehere}[1]{\iflatexml\space(#1)\else\protect\footnotemark\fi}
\DeclareRobustCommand{\notetext}[1]{\iflatexml\else\footnotetext{#1}\fi}

\usepackage{listings}
\usepackage{hyperref} %
\usepackage[nameinlink]{cleveref} %
\newcounter{qexample} %
\crefname{qexample}{Example}{Examples}
\Crefname{qexample}{Example}{Examples}

\usepackage{float}
\usepackage{algorithm}
\usepackage{algpseudocode}
\usepackage{needspace} %

\definecolor{rwth}   {RGB}{  0  84 159}
\definecolor{rwth-75}{RGB}{ 64 127 183}
\definecolor{rwth-50}{RGB}{142 186 229}
\definecolor{rwth-25}{RGB}{199 221 242}
\definecolor{rwth-10}{RGB}{232 241 250}

\definecolor{black}   {RGB}{  0   0   0}
\definecolor{black-75}{RGB}{100 101 103}
\definecolor{black-50}{RGB}{156 158 159}
\definecolor{black-25}{RGB}{207 209 210}
\definecolor{black-10}{RGB}{236 237 237}

\definecolor{magenta}   {RGB}{227   0 102}
\definecolor{magenta-75}{RGB}{233  96 136}
\definecolor{magenta-50}{RGB}{241 158 177}
\definecolor{magenta-25}{RGB}{249 210 218}
\definecolor{magenta-10}{RGB}{253 238 240}

\definecolor{yellow}   {RGB}{255 237   0}
\definecolor{yellow-75}{RGB}{255 240  85}
\definecolor{yellow-50}{RGB}{255 245 155}
\definecolor{yellow-25}{RGB}{255 250 209}
\definecolor{yellow-10}{RGB}{255 253 238}

\definecolor{petrol}   {RGB}{  0  97 101}
\definecolor{petrol-75}{RGB}{ 45 127 131}
\definecolor{petrol-50}{RGB}{125 164 167}
\definecolor{petrol-25}{RGB}{191 208 209}
\definecolor{petrol-10}{RGB}{230 236 236}

\definecolor{turkis}   {RGB}{  0 152 161}
\definecolor{turkis-75}{RGB}{  0 177 183}
\definecolor{turkis-50}{RGB}{137 204 207}
\definecolor{turkis-25}{RGB}{202 231 231}
\definecolor{turkis-10}{RGB}{235 246 246}

\definecolor{grun}   {RGB}{ 87 171  39}
\definecolor{grun-75}{RGB}{141 192  96}
\definecolor{grun-50}{RGB}{184 214 152}
\definecolor{grun-25}{RGB}{221 235 206}
\definecolor{grun-10}{RGB}{242 247 236}

\definecolor{maigrun}   {RGB}{189 205   0}
\definecolor{maigrun-75}{RGB}{208 217  92}
\definecolor{maigrun-50}{RGB}{224 230 154}
\definecolor{maigrun-25}{RGB}{240 243 208}
\definecolor{maigrun-10}{RGB}{249 250 237}

\definecolor{orange}   {RGB}{246 168   0}
\definecolor{orange-75}{RGB}{250 190  80}
\definecolor{orange-50}{RGB}{253 212 143}
\definecolor{orange-25}{RGB}{254 234 201}
\definecolor{orange-10}{RGB}{255 247 234}

\definecolor{rot}   {RGB}{204   7  30}
\definecolor{rot-75}{RGB}{216  92  65}
\definecolor{rot-50}{RGB}{230 150 121}
\definecolor{rot-25}{RGB}{243 205 187}
\definecolor{rot-10}{RGB}{250 235 227}

\definecolor{bordeaux}   {RGB}{161  16  53}
\definecolor{bordeaux-75}{RGB}{182  82  86}
\definecolor{bordeaux-50}{RGB}{205 139 135}
\definecolor{bordeaux-25}{RGB}{229 197 192}
\definecolor{bordeaux-10}{RGB}{245 232 229}

\definecolor{violett}   {RGB}{ 97  33  88}
\definecolor{violett-75}{RGB}{131  78 117}
\definecolor{violett-50}{RGB}{168 133 158}
\definecolor{violett-25}{RGB}{210 192 205}
\definecolor{violett-10}{RGB}{237 229 234}

\definecolor{lila}   {RGB}{122 111 172}
\definecolor{lila-75}{RGB}{155 145 193}
\definecolor{lila-50}{RGB}{188 181 215}
\definecolor{lila-25}{RGB}{222 218 235}
\definecolor{lila-10}{RGB}{242 240 247}

\usepgfplotslibrary{groupplots}
\definecolor{tobase}{HTML}{8A8C8E} %
\colorlet{toer}{rwth}         %
\colorlet{torr}{magenta}      %
\DeclareRobustCommand{\unitcut}{UNIT$_{cut}$}
\pgfplotsset{
  tradeoff/.style={
    scale only axis,
    grid=major,
    grid style={black-10, line width=0.3pt},
    axis line style={black-50, line width=0.3pt},
    tick style={black-50, line width=0.3pt},
    tick label style={font=\scriptsize},
    label style={font=\scriptsize},
    title style={font=\scriptsize, yshift=-4pt},
    xlabel style={yshift=2pt},
    ylabel style={yshift=-2pt},
    clip mode=individual,
  },
  tolabel/.style={
    nodes near coords,
    point meta=explicit symbolic,
    visualization depends on={value \thisrow{anchor} \as \toanchor},
    every node near coord/.append style={anchor=\toanchor, font=\tiny, inner sep=1.2pt, outer sep=3pt, text=black},
  },
  pt/base/.style={only marks, mark=*, mark size=1.9pt, tobase},
  pt/er/.style={only marks, mark=square*, mark size=1.9pt, toer},
  pt/rr/.style={only marks, mark=diamond*, mark size=2.9pt, torr},
  pt/rrabl/.style={mark=diamond, mark size=2.9pt, torr, line width=0.5pt, mark options={solid, line width=0.6pt}},
  pt/rrfilt/.style={only marks, mark=triangle, mark size=2.7pt, torr, mark options={line width=0.6pt}},
  isof/.style={black-25, line width=0.4pt, samples=80, forget plot},
  tolegend/.style={
    legend columns=-1,
    legend style={font=\scriptsize, draw=none, fill=none,
      /tikz/every even column/.append style={column sep=0.35cm}},
    legend image post style={mark size=2.2pt},
  },
}
\tikzset{isolabel/.style={font=\tiny, text=black-50, inner sep=1pt}}

\defcitealias{gpt4o}{OpenAI, 2024}
\defcitealias{olmo2026olmo3}{Team OLMo, 2025}

\newcommand\blfootnote[1]{%
  \begingroup
  \renewcommand\thefootnote{}\footnote{#1}%
  \addtocounter{footnote}{-1}%
  \endgroup
}
\DeclareRobustCommand{\qwenmodel}{Qwen~3~4B}
\DeclareRobustCommand{\qweninstruct}{Qwen~3 Instruct}
\DeclareRobustCommand{\qwenbaseid}{\texttt{Qwen3-4B-Base}}
\DeclareRobustCommand{\qweninstructid}{\texttt{Qwen/Qwen3-4B-Instruct-2507}}
\DeclareRobustCommand{\olmomodel}{OLMo~3~7B}
\DeclareRobustCommand{\olmoinstruct}{OLMo~3 Instruct}
\DeclareRobustCommand{\olmobaseid}{\texttt{OLMo-3-1025-7B}}
\DeclareRobustCommand{\tulusft}{T\"ulu~3 SFT}
\DeclareRobustCommand{\gptfour}{\texttt{gpt-4}}
\DeclareRobustCommand{\gptfouromini}{\texttt{gpt-4o-mini}}
\DeclareRobustCommand{\gptfivemini}{\texttt{gpt-5-mini}}
\DeclareRobustCommand{\gptfiveminifull}{\texttt{gpt-5-mini-2025-08-07}}

\title{Stick to What You Know:\\A Study of Knowledge-Aligned Supervised Fine-Tuning}

\iflatexml
\author{%
  Arthur Becker\\AppTek GmbH, Aachen, Germany \& F-Bureaucracy UG, Aachen, Germany\\Equal contribution.
  \and Jakob Kemmler\\AppTek GmbH, Aachen, Germany\\Equal contribution.
  \and David Thulke\\AppTek GmbH, Aachen, Germany \& RWTH Aachen University, Germany\\Corresponding author: \texttt{dthulke@apptek.com}
  \and Christine Sch\"afer\\AppTek GmbH, Aachen, Germany
  \and Christian Dugast\\AppTek GmbH, Aachen, Germany
  \and Hermann Ney\\AppTek GmbH, Aachen, Germany \& RWTH Aachen University, Germany}
\else
\author{Arthur Becker\textsuperscript{*,1,2} \, Jakob Kemmler\textsuperscript{*,1} \, David Thulke\textsuperscript{\textdaggerdbl,1,3} \\ \textbf{Christine Schäfer\textsuperscript{1}} \, \textbf{Christian Dugast\textsuperscript{1}} \, \textbf{Hermann Ney\textsuperscript{1,3}} \\ \\
        \textsuperscript{1}AppTek GmbH, Aachen, Germany\\
        \textsuperscript{2}F-Bureaucracy UG, Aachen, Germany\\
        \textsuperscript{3}Machine Learning and Human Language Technology, RWTH Aachen University, Germany}
\fi

\begin{document}
\maketitle
\iflatexml\setcounter{footnote}{0}\fi
\begin{abstract}
Supervised fine-tuning (SFT) trains a base language model to imitate target responses, and these targets may require knowledge the base model has not robustly internalized.
We study this as a source of hallucinations and frame a group of mitigation methods as \emph{knowledge-aligned SFT}: constraining SFT training targets to the base model's parametric knowledge.
Under a unified setup, we compare existing generation-based and estimation-based knowledge-alignment methods and introduce two new variants: \emph{Evidence Rewrite}, which verifies base-model generations using external evidence, and \emph{Recall Rewrite}, which retains claims only when they can be consistently recalled by the base model.
Experiments with \qwenmodel{} and \olmomodel{} show that knowledge-aligned SFT can reduce factual hallucinations on WildHalu and Biography while largely preserving general capabilities.
\emph{Recall Rewrite} yields the strongest factuality gains and improves refusal behavior on UnknownBench.
It thereby confirms that SFT targets beyond the base model's knowledge drive hallucination behavior.
\end{abstract}%

\section{Introduction}%
\iflatexml\else%
\makeatletter%
\ifacl@anonymize\else%
  \blfootnote{\textsuperscript{*} Equal contribution.}%
  \blfootnote{\textsuperscript{\textdaggerdbl} Corresponding author: \texttt{dthulke@apptek.com}}%
\fi%
\makeatother%
\fi%
Large language models (LLMs) often generate fluent answers that contain unsupported or false claims, a failure mode commonly referred to as hallucination.
We focus on \emph{factuality hallucinations}: generated claims that contradict real-world facts \cite{survey-hallucinations}.
This is especially problematic for instruction-following models, where users frequently rely on detailed answers in domains where errors are difficult to detect.

\begin{figure}[t]
\centering

\colorlet{pkborder}{rwth}
\colorlet{pkfill}{rwth-10}
\colorlet{gborder}{turkis}
\colorlet{gfill}{turkis-10}
\colorlet{wkborder}{black-50}
\colorlet{wkfill}{black-10}
\colorlet{dvborder}{rot}
\colorlet{dvfill}{rot-10}
\colorlet{daborder}{grun}
\colorlet{dafill}{grun-10}
\colorlet{hzlabel}{rot}

\iflatexml
\htmlfig{knowledge_alignment_figure}{Venn diagrams contrasting standard SFT with knowledge-aligned SFT. Top: dataset knowledge extends beyond the base model knowledge, so the fine-tuned generation space spills outside world knowledge, marked as the hatched hallucination zone. Bottom: knowledge-aligned SFT keeps the constructed dataset inside the base model knowledge, shrinking that zone.}%
\else
\begin{tikzpicture}[
  every node/.style={font=\small},
]

\def\bigR{1.2}
\def\medR{1.0}
\def\smallR{0.68}
\def\rowsep{3.3}

\def\leftCX{1.6}
\def\rightCX{6.0}
\def\arrowX{3.9}

\def\topY{0}
\pgfmathsetmacro{\botY}{-\rowsep}

\node[font=\bfseries\small, anchor=west] at (0, \topY+\bigR+0.4)
  {Standard SFT};

\fill[pkfill] (\leftCX, \topY) circle (\bigR);
\draw[pkborder, line width=1.2pt] (\leftCX, \topY) circle (\bigR);

\fill[dvfill] (\leftCX+1.0, \topY) circle (\smallR);
\draw[dvborder, line width=1.2pt] (\leftCX+1.0, \topY) circle (\smallR);

\node[font=\scriptsize, pkborder!85!black]
  at (\leftCX-0.6, \topY) {$\mathcal{K}\!(\!M_\text{base}\!)$};
\node[font=\scriptsize, dvborder]
  at (\leftCX+1.0, \topY) {$\mathcal{K}(\mathcal{D})$};

\draw[->, >=stealth, line width=2.0pt, black!35]
  (\arrowX-0.3, \topY) -- (\arrowX+0.3, \topY);
\node[above=3pt, font=\scriptsize\bfseries, black!45] at (\arrowX, \topY)
  {SFT};

\fill[wkfill] (\rightCX+0.55, \topY) circle (\bigR);
\fill[gfill] (\rightCX-0.55, \topY) circle (\medR);

\begin{scope}
  \clip (\rightCX-0.55, \topY) circle (\medR);
  \fill[pattern=north east lines, pattern color=rot-75]
    (\rightCX-3, \topY-3) rectangle (\rightCX+3, \topY+3);
  \fill[gfill] (\rightCX+0.55, \topY) circle (\bigR);
\end{scope}

\draw[gborder, line width=1.2pt] (\rightCX-0.55, \topY) circle (\medR);
\draw[wkborder, line width=1.2pt, dashed] (\rightCX+0.55, \topY) circle (\bigR);

\node[font=\scriptsize, gborder!90!black]
  at (\rightCX-0.1, \topY) {{$\mathcal{G}\!(\!M_\text{SFT}\!)$}};
\node[font=\scriptsize, wkborder!85!black]
  at (\rightCX+1.45, \topY) {$\mathcal{W}$};

\node[font=\bfseries\small, anchor=west] at (0, \botY+\bigR+0.4)
  {Knowledge-Aligned SFT};

\fill[pkfill] (\leftCX, \botY) circle (\bigR);
\draw[pkborder, line width=1.2pt] (\leftCX, \botY) circle (\bigR);

\fill[dafill] (\leftCX+0.55, \botY) circle ({\smallR-0.05});
\draw[daborder, line width=1.2pt] (\leftCX+0.55, \botY) circle ({\smallR-0.05});

\node[font=\scriptsize, pkborder!85!black]
  at (\leftCX-0.6, \botY) {$\mathcal{K}\!(\!M_\text{base}\!)$};
\node[font=\scriptsize, daborder]
  at (\leftCX+0.6, \botY) {$\mathcal{K}(\mathcal{D}^*)$};

\draw[->, >=stealth, line width=2.0pt, black!35]
  (\arrowX-0.3, \botY) -- (\arrowX+0.3, \botY);
\node[above=3pt, font=\scriptsize\bfseries, black!45] at (\arrowX, \botY)
  {SFT};

\fill[wkfill] (\rightCX+0.55, \botY) circle (\bigR);
\fill[gfill] (\rightCX+0.2, \botY) circle (\medR);

\begin{scope}
  \clip (\rightCX+0.2, \botY) circle (\medR);
  \fill[pattern=north east lines, pattern color=rot-75]
    (\rightCX-2, \botY-3) rectangle (\rightCX+3, \botY+3);
  \fill[gfill] (\rightCX+0.55, \botY) circle (\bigR);
\end{scope}

\draw[gborder, line width=1.2pt] (\rightCX+0.2, \botY) circle (\medR);
\draw[wkborder, line width=1.2pt, dashed] (\rightCX+0.55, \botY) circle (\bigR);

\node[font=\scriptsize, gborder!90!black]
  at (\rightCX+0.2, \botY) {$\mathcal{G}\!(\!M_\text{SFT}\!)$};
\node[font=\scriptsize, wkborder!85!black]
  at (\rightCX+1.45, \botY) {$\mathcal{W}$};

\draw[->, >=stealth, hzlabel, line width=0.6pt]
  (\rightCX-1.1, \botY+1.1) -- (\rightCX-0.72, \botY+0.02);
\draw[->, >=stealth, hzlabel, line width=0.6pt]
  (\rightCX-1.1, \botY+1.8) -- (\rightCX-1.0, \topY-0.4);
\node[font=\scriptsize\bfseries, hzlabel, anchor=south, align=center]
  at (\rightCX-1.1, \botY+1.1)
  {Hallucination\\[-1pt]Zone};

\end{tikzpicture}
\fi
\caption{%
  \textbf{Left:} comparing the dataset knowledge \textcolor{dvborder}{$\mathcal{K}(\mathcal{D})$} used for SFT with the base model's parametric knowledge \textcolor{pkborder}{$\mathcal{K}(M_{\text{base}})$};
  \textbf{Right:} the resulting generation space \textcolor{gborder}{$\mathcal{G}(M_{\text{SFT}})$} relative to world knowledge \textcolor{wkborder}{$\mathcal{W}$}.
  The hatched area marks the \emph{hallucination zone}: generations outside \textcolor{wkborder}{$\mathcal{W}$}.
  \textbf{Top:} Standard SFT on \textcolor{dvborder}{$\mathcal{D}$} includes claims beyond \textcolor{pkborder}{$\mathcal{K}(M_{\text{base}})$}, encouraging generation beyond what the model knows.
  \textbf{Bottom:} Knowledge-aligned SFT constructs \textcolor{daborder}{$\mathcal{D}^*$} within \textcolor{pkborder}{$\mathcal{K}(M_{\text{base}})$}, reducing the part of \textcolor{gborder}{$\mathcal{G}(M_{\text{SFT}})$} outside \textcolor{wkborder}{$\mathcal{W}$}.%
}%
\label{fig:knowledge_alignment}%
\end{figure}
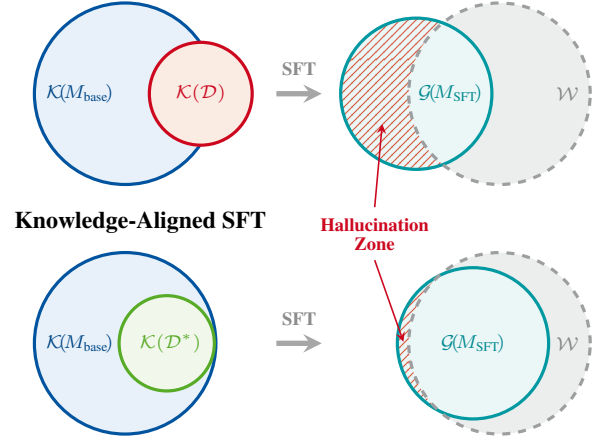

Supervised fine-tuning (SFT) is a central step for turning a pre-trained base model into an instruction-following model \cite{ouyang2022training}.
Instruction tuning exposes models to broad task mixtures to improve instruction following and generalization to new tasks and domains \cite{wei2022finetuned, longpre2023flan}.
However, factual SFT examples provide two intertwined signals: they teach the model \emph{how} to answer, while also requiring it to produce factual claims. %
If these claims are not robustly internalized during pre-training, models are pushed to produce plausible claims beyond their parametric knowledge.
Prior work established that this is a driver of hallucinations in SFT \cite{gekhman-etal-2024-fine, flame, wu2025balancingtruthfulnessinformativenessuncertaintyaware, kaplan2026finetuningencourageshallucinationsfix}.
More broadly, SFT is a poor vehicle for injecting new knowledge, because learning new factual associations reliably requires extensive evidence \cite{kandpal2023large} and the SFT stage is comparatively small (\citealp{Guo_2025}; \citetalias{olmo2026olmo3}).
Retrieval augmentation or continued pre-training are better suited for this purpose \cite{ovadia-etal-2024-fine, xu-etal-2023-kilm}.

Additionally, SFT strongly shapes the model's response behavior:
it teaches when to answer, how much detail to provide, and how confidently to present claims.
If supervision targets require knowledge outside the base model's parameters, SFT rewards valid-looking answers under uncertainty rather than respect for the model's knowledge boundary.
This encourages guessing rather than refusing, echoing the argument that common training and evaluation setups reward answers under uncertainty while penalizing expressions of uncertainty \cite{kalai2025language}.

We study this problem through the lens of \emph{knowledge alignment}: matching SFT targets to the factual knowledge already available in the base model.
Knowledge-aligned SFT therefore aims to preserve SFT's behavioral benefits while reducing the incentive to guess beyond the base model's knowledge.
\Cref{fig:knowledge_alignment} summarizes our framing.
Standard SFT may require the model to imitate responses containing facts it does not know, which expands the space of plausible-looking but unsupported generations.
Knowledge-aligned SFT instead constrains factual supervision to what the base model can support.

In this work, we provide a controlled comparison of knowledge-aligned SFT methods under a unified framework.
We compare generation-based alignment, represented by FLAME \cite{flame}, with estimation-based alignment, represented by uncertainty-aware filtering UNIT$_{cut}$ \cite{wu2025balancingtruthfulnessinformativenessuncertaintyaware}, and introduce two variants: \emph{Evidence Rewrite}, which filters base-model generations through external claim verification, and \emph{Recall Rewrite}, which probes whether claims can be consistently recalled by the base model.
Across factuality, refusal, and general capability evaluations, knowledge-aligned supervision reduces measured hallucinations while preserving broad model capabilities.
Our contribution is to operationalize this premise on real instruction-tuning data.
We classify each individual claim of a training response as known or unknown to the base model and intervene on the SFT targets accordingly.
In \Cref{subsec:cr_ablation} we additionally reproduce the causal effect within our pipeline by varying only the share of known claims in otherwise identical training data.
We release the Recall Rewrite training data for both base models together with all intermediate pipeline outputs (\Cref{subsec:experimental_settings}, \Cref{fn:data-release}).

\section{Framework}
\label{sec:framework}

We now formalize knowledge-aligned SFT. Let $\mathcal{W}$ denote real-world knowledge, and let $\mathcal{K}(M_{\text{base}})$ denote the parametric knowledge of a base model $M_{\text{base}}$.
This knowledge is inherently incomplete: it reflects what the model encountered and robustly internalized during pre-training, not the totality of $\mathcal{W}$.
Following \Cref{fig:knowledge_alignment}, let $\mathcal{G}(M)$ denote the set of factual claims a model may generate after instruction tuning.
We define the \emph{hallucination zone} as $\mathcal{G}(M) \setminus \mathcal{W}$, i.e., generations that fall outside real-world knowledge.
Our goal is not to make SFT inject new factual knowledge, but to reduce the pressure for $M_{\text{SFT}}$ to generate claims unsupported by $M_{\text{base}}$, thereby shrinking the hallucination zone.

Let $\mathcal{D} = \{(P, R)\}$ be an SFT dataset of prompt-response pairs. We decompose each response $R$ into a set of atomic claims $\mathcal{C}(R \mid P) = \{c_1, \dots, c_N\}$, restricting to claims that are (i) factual, procedural, or structural, and (ii) not already provided in $P$, i.e., claims the model must supply from its own parametric knowledge.
Importantly, this decomposition includes \textit{meta-knowledge} that is implicit but necessary for constructing $R$ under $P$.
For instance, to \textit{"Write a haiku about..."}, one requires knowledge of the form's three-line structure and syllabic constraints.
We write $\mathcal{K}(\mathcal{D}) = \bigcup_{(P,R) \in \mathcal{D}} \mathcal{C}(R \mid P)$ for the knowledge contained in the dataset under this claim decomposition.
We say a claim $c$ is \emph{known} to $M_\text{base}$ if $c \in \mathcal{K}(M_\text{base})$ and \emph{unknown} otherwise.
\Cref{sec:glossary} collects this and the related terminology used throughout the paper.
A training example is knowledge-aligned iff all its claims are known.
The objective is therefore to construct $\mathcal{D}^*$ such that $\mathcal{C}(R^* \mid P) \subseteq \mathcal{K}(M_{\text{base}})$ for all $(P, R^*) \in \mathcal{D}^*$.
\emph{Knowledge-aligned SFT} is any procedure that constructs such a $\mathcal{D}^*$ from $\mathcal{D}$ and fine-tunes $M_\text{base}$ on it.
It addresses one specific subset of the broader SFT alignment problem: the factual knowledge that supervision requires, rather than mismatches in style, skills, or out-of-distribution behavior.

Since $\mathcal{K}(M_{\text{base}})$ is latent, it must be approximated.
Each method therefore comes with its own way of classifying claims as known or unknown.
We distinguish two strategies.

\paragraph{Knowledge Alignment via Generation.}
One approach constructs supervision from the model's own outputs:
a response $\hat{R} \sim M_{\text{base}}$ replaces the gold target $R$, on the assumption that self-generated content is implicitly constrained to $\mathcal{K}(M_{\text{base}})$.
This assumption is imperfect in two ways.
$M_{\text{base}}$ may hallucinate, so $\hat{R}$ can contain claims that are not knowledge-aligned.
And $\mathcal{K}(M_{\text{base}})$ may contain false beliefs, so even knowledge-aligned claims in $\hat{R}$ can be factually wrong.

\paragraph{Knowledge Alignment via Estimation.}
An alternative approach estimates, for each claim $c \in \mathcal{C}(R \mid P)$, whether it lies within $\mathcal{K}(M_{\text{base}})$, and removes or modifies those that do not.
In practice, this relies on heuristic proxy signals such as model confidence \cite{wu2025balancingtruthfulnessinformativenessuncertaintyaware}.

\section{Methods}

We instantiate both alignment strategies from \Cref{sec:framework} with concrete methods and propose improved variants.

\subsection{Existing Work}
\label{subsec:existing_work}

We provide a short overview of existing methods. Further details are provided in \Cref{sec:baseline_method_details}.

\paragraph{Alignment via Generation (FLAME).}
\citet{flame} present \emph{FLAME}, which replaces gold responses $R$ with outputs $\hat{R} \sim M_{\text{base}}$, implicitly assuming self-generated content is constrained to $\mathcal{K}(M_{\text{base}})$.
For prompts that are not \emph{knowledge-seeking}, i.e., that can be answered without factual knowledge (e.g., summarization), the gold response is retained.
Under our framework, FLAME implicitly classifies every self-generated claim in $\mathcal{C}(\hat{R} \mid P)$ as known.
$\hat{R}$ may contain factually incorrect claims, and training on these may reinforce rather than reduce hallucination behavior.
Furthermore, as $\hat{R}$ replaces $R$ entirely, claims within $\mathcal{K}(M_{\text{base}})$ that are present in $R$ but not generated are lost.

\paragraph{Alignment via Estimation (UNIT$_{cut}$).}
\citet{wu2025balancingtruthfulnessinformativenessuncertaintyaware} suggest \emph{UNIT$_{cut}$}, which filters atomic claims from each response $R$ based on their \emph{claim-conditioned probability} (CCP): the likelihood the model assigns to a claim given its context.
UNIT$_{cut}$ thus classifies a claim $c \in \mathcal{C}(R \mid P)$ as known iff its CCP exceeds a threshold, and $R^*$ retains only these claims.
While computationally efficient, CCP is a token-level signal sensitive to phrasing and position, and confidence-based truthfulness estimation has been found to underperform reference-based verification \cite{facttune}.

\subsection{Proposed Methods}

Both proposed methods, as well as the two baselines above, are instances of a single data-construction procedure that differs only in the source response and in how claims are classified as known.
We make this common structure explicit in \Cref{sec:unified_algorithm}.

\subsubsection{Evidence Rewrite}
\label{subsec:factuality_rewrite}

\begin{figure}
    \centering
\iflatexml
\htmlfig{factuality-rewrite}{Flow diagram of Evidence Rewrite. Knowledge-seeking prompts are answered by the base model, decomposed into claims, fact-checked against retrieved Wikipedia evidence, filtered to supported claims and rewritten; non-knowledge-seeking prompts keep the gold response. An inset expands the fact-checking step.}%
\else
    \scalebox{\rewritefigurescale}{
    \begin{tikzpicture}[
        >=Latex, 
        font=\sffamily\small,
        node distance=1.5cm and 0cm,
        base/.style={
            draw, rectangle, 
            minimum width=4.5cm, minimum height=1.1cm, 
            text centered, line width=0.8pt, rounded corners=2pt
        },
        prompt_style/.style={base, fill=rwth-10, draw=rwth, text=rwth},
        response_style/.style={base, fill=petrol-10, draw=petrol, text=petrol},
        claim_box/.style={
            draw, rectangle, 
            minimum width=1.5cm, minimum height=0.5cm, 
            text centered, font=\sffamily\small, line width=0.5pt
        },
        stack_container/.style={draw, dashed, inner sep=10pt, rounded corners=3pt, color=gray!50},
        inner_icon/.style={
            circle, draw, inner sep=0pt, 
            font=\sffamily\large\bfseries, 
            line width=1.5pt, 
            minimum size=18pt
        }
    ]

    \node[prompt_style] (prompt) at (0, 5.1) {\hspace{15pt}Prompt $P$};
    \node[inner_icon, draw=rwth, text=rwth, anchor=west, xshift=6pt] at (prompt.west) {?};

    \coordinate (branch_center) at (0, 4.0);
    \draw[line width=1.0pt] (prompt.south) -- (branch_center);
    
    \draw[line width=1.0pt] (-2.8, 4.0) -- (2.8, 4.0);

    \node[response_style] (flame) at (-2.8, 2.5) {\hspace{25pt}FLAME response $\hat{R}$};
    \node[inner_icon, draw=petrol, text=petrol, anchor=west, xshift=6pt] at (flame.west) {!};
    
    \draw[->, line width=1.0pt] (-2.8, 4.0) --
        node[pos=0, anchor=south, font=\sffamily\small\itshape]
        {knowledge-seeking}
        node[fill=white, inner sep=2pt, pos=0.5] {sample from $M_\text{base}$} 
        (flame.north);

    \node[claim_box, fill=grun-25, draw=grun] (c1_stack4) [below=1.7cm of flame, xshift=5pt, yshift=-5pt] {};
    \node[claim_box, fill=grun-25, draw=grun] (c1_stack3) at ([shift={(-2pt,2pt)}]c1_stack4) {};
    \node[claim_box, fill=rot-25, draw=rot] (c1_stack2) at ([shift={(-2pt,2pt)}]c1_stack3) {};
    \node[claim_box, fill=grun-25, draw=grun]   (c1_stack1) at ([shift={(-2pt,2pt)}]c1_stack2) {};
    \node[claim_box, fill=rot-25, draw=rot]   (c1_main)   at ([shift={(-2pt,2pt)}]c1_stack1) {Claim $c_1$};
    
    \node[stack_container, fit=(c1_stack3) (c1_main)] (stack1_box) {};

    \node[claim_box, fill=grun-25, draw=grun] (c2_stack2) [below=1.5cm of stack1_box, xshift=2pt, yshift=-3pt] {};
    \node[claim_box, fill=grun-25, draw=grun] (c2_stack1) at ([shift={(-2pt,2pt)}]c2_stack2) {};
    \node[claim_box, fill=grun-25, draw=grun] (c2_main)   at ([shift={(-2pt,2pt)}]c2_stack1) {Claim $c_2$};

    \node[stack_container, fit=(c2_stack2) (c2_main)] (stack2_box) {};

    \node[response_style, below=1.1cm of stack2_box] (final) {\hspace{25pt}Rewritten response $R^*$};
    \node[inner_icon, draw=petrol, text=petrol, anchor=west, xshift=6pt] at (final.west) {!};

    \path (flame.south) -- (stack1_box.north) coordinate[midway] (zoom_pos);
    \draw[->, line width=0.6pt] (flame.south) -- 
        node[left=6pt, font=\itshape\small, midway, align=right] {fact checking}
        (stack1_box.north);

    \node[circle, draw=petrol, text=petrol, line width=1pt, minimum size=15pt,
          inner sep=0pt, fill=petrol-10, font=\sffamily\scriptsize\bfseries]
          (zoom_circle) at (zoom_pos) {D};

    \draw[->, line width=0.6pt] (stack1_box.south) -- 
        node[left=6pt, font=\itshape\small, midway, align=right] {filtering out\\\textcolor{rot}{unsupported}}
        (stack2_box.north);

    \draw[->, line width=0.6pt] (stack2_box.south) -- 
        node[left=6pt, font=\itshape\small, midway, align=right] {summarizing\\with $M_\text{rewriter}$}
        (final.north);

    \node[response_style, at={(2.8, 2.5)}] (gold) {\hspace{25pt}Gold response $R$};
    \node[inner_icon, draw=petrol, text=petrol, anchor=west, xshift=6pt] at (gold.west) {!};
    
    \draw[->, line width=1.0pt] (2.8, 4.0) --
        node[pos=0, anchor=south, font=\sffamily\small\itshape]
        {\textbf{non}-knowledge-seeking}
        (gold.north);

    \node[draw, rectangle, minimum width=5.6cm, minimum height=3.5cm,
          inner xsep=0.2cm, inner ysep=6pt, line width=0.8pt, fill=white,
          rounded corners=5pt, right=2.2cm of zoom_circle, yshift=-3.1cm] (detail_box) {
        \begin{minipage}[t]{5.2cm}
            \centering
            \begin{tikzpicture}[>=Latex, font=\sffamily\small]
                \node[claim_box, rectangle, fill=petrol-10, draw=petrol, text=petrol, rounded corners=0] (ci) {Response $\hat{R}$};

                \node[rectangle, minimum width=0.3cm, minimum height=1.1cm,
                      inner xsep=0pt, text centered, below=0.6cm of ci]
                      (placeholder_q) {...};
                \node[claim_box, rectangle, fill=rot-25, draw=rot, rounded corners=0,
                      left=0.05cm of placeholder_q, minimum width=2.4cm,
                      inner xsep=2pt] (q1) {Claim $c_1$};
                \node[claim_box, rectangle, fill=grun-25, draw=grun, rounded corners=0,
                      right=0.05cm of placeholder_q, minimum width=2.4cm,
                      inner xsep=2pt] (qm) {Claim $c_n$};

                \node[rectangle, minimum width=0.3cm, minimum height=1.1cm,
                      inner xsep=0pt, text centered, below=0.4cm of placeholder_q]
                      (placeholder_e) {...};
                \node[claim_box, rectangle, fill=lila-10, draw=lila, rounded corners=0,
                      left=0.05cm of placeholder_e, minimum width=2.4cm,
                      inner xsep=2pt, text=lila] (a1) {Evidence $e_{1,1}$};
                \node[claim_box, rectangle, fill=lila-10, draw=lila, rounded corners=0,
                      right=0.05cm of placeholder_e, minimum width=2.4cm,
                      inner xsep=2pt, text=lila] (ar) {Evidence $e_{1,5}$};

                \node[ rectangle, fill=white, draw=black, rounded corners=0,
                 minimum height=0.8cm, below=0.27cm of placeholder_e, minimum width=4cm, text=black] (mverifier) {$M_\text{verifier}$};

                \draw[->] (ci.south) -- (q1.north);
                \draw[->] (ci.south) -- (qm.north)
                    node[midway, right, overlay, font=\small, xshift=-1.7cm] {decompose};
                \draw[->] (q1.south) -- (a1.north);
                \draw[->] (q1.south) -- (ar.north)
                    node[midway, right, overlay, font=\small, xshift=-0.7cm,
                         text width=2cm] {retrieve\\from Wiki};

                \draw[->, dashed] (a1.south) -- (mverifier);
                \draw[->, dashed] (ar.south) -- (mverifier);
            \end{tikzpicture}
            
            {
                \small 
                \sffamily  
                \raggedright
                $M_\text{verifier}$ decides whether the claim is \textcolor{grun}{supported} or \textcolor{rot}{unsupported}.
            }
        \end{minipage}
    };
    \node[font=\itshape\small, below=2pt of detail_box.south]
        {\textcolor{petrol}{\textsf{\bfseries\upshape D}}\enspace Fact Checking (expanded)};

    \end{tikzpicture}
    }
    \vspace*{-0.05cm}
\fi
    \caption{Evidence Rewrite. For knowledge-seeking prompts, a base-model response $\hat{R}$ is decomposed, verified against retrieved evidence, filtered to supported claims, and rewritten as $R^*$.}
    \label{fig:factuality-rewrite-illustration}
\end{figure}
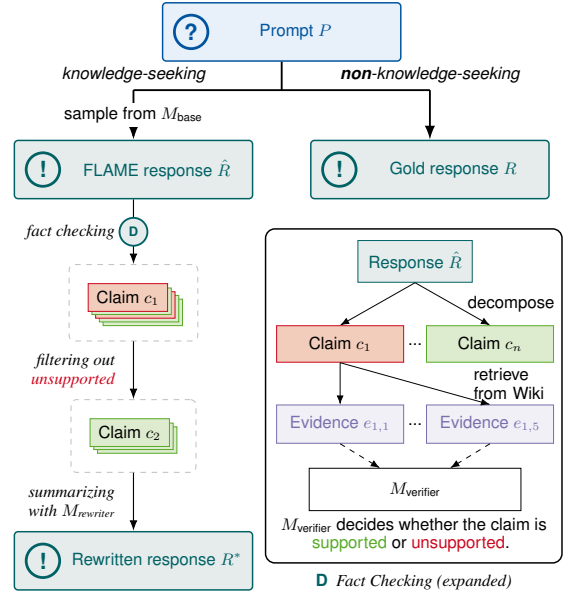

\emph{Evidence Rewrite} (\Cref{fig:factuality-rewrite-illustration}) refines the generation-based alignment strategy of FLAME by adding an external verification step.
FLAME treats every self-generated claim as known even though it may be factually incorrect, so we filter out claims that are not supported by external evidence.

As in FLAME, prompts that are not knowledge-seeking keep their gold response.
For each knowledge-seeking prompt $P$, we sample a long-form response $\hat{R} \sim M_{\text{base}}$.
We apply a standard fact-checking pipeline (claim decomposition, evidence retrieval, and verification; \Cref{subsec:fact-checking}) to classify each claim in $\hat{R}$ as supported or unsupported.
We then use a rewriter model $M_{\text{rewriter}}$, conditioned on the prompt and the supported claims, to compose a fluent aligned response $R^*$.
Under our framework, Evidence Rewrite thus classifies a claim as known iff it is self-generated and verified as supported.
If the supported claims provide insufficient information to address $P$, the prompt instructs $M_\text{rewriter}$ to refuse to respond instead.
To mitigate response shortening caused by claim filtering, we introduce an additional brainstorming step before fact-checking that instructs the base model to generate a longer, more detailed response given an initial short generation (see \Cref{fig:brainstorm_example}).

\subsubsection{Recall Rewrite}
\label{subsec:confidence_rewrite}

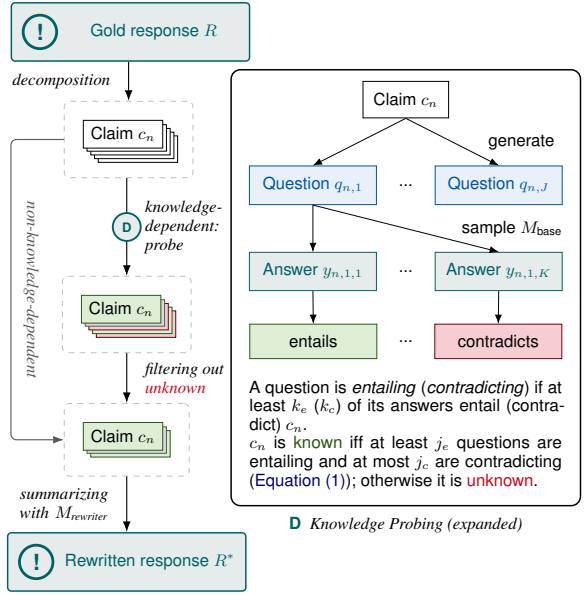
\begin{figure}
    \centering
\iflatexml
\htmlfig{confidence-rewrite}{Flow diagram of Recall Rewrite. Each knowledge-dependent claim of the gold response is turned into probing questions, answered by the base model and checked for entailment; claims that are not consistently recalled are filtered out and the rest rewritten. An inset expands the knowledge-probing step.}%
\else
    \makebox[\linewidth][c]{\hspace*{-0.05cm}\scalebox{\rewritefigurescale}{
    \begin{tikzpicture}[
        >=Latex, 
        font=\sffamily\small,
        node distance=1.5cm and 0cm,
        base/.style={
            draw, rectangle, 
            minimum width=4.5cm, minimum height=1.1cm, 
            text centered, line width=0.8pt, rounded corners=2pt
        },
        prompt_style/.style={base, fill=rwth-10, draw=rwth, text=rwth},
        response_style/.style={base, fill=petrol-10, draw=petrol, text=petrol},
        claim_box/.style={
            draw, rectangle, 
            minimum width=1.5cm, minimum height=0.5cm, 
            text centered, font=\sffamily\small, line width=0.5pt
        },
        stack_container/.style={draw, dashed, inner sep=10pt, rounded corners=3pt, color=gray!50},
        inner_icon/.style={
            circle, draw, inner sep=0pt, 
            font=\sffamily\large\bfseries, 
            line width=1.5pt, 
            minimum size=18pt
        }
    ]

    \node[response_style] (flame)  {\hspace{25pt}Gold response $R$};
    \node[inner_icon, draw=petrol, text=petrol, anchor=west, xshift=6pt] at (flame.west) {!};

    \node[claim_box, fill=white, draw=black] (c0_stack4) at ([shift={(4pt,-1.65cm)}]flame.south) {};
    \node[claim_box, fill=white, draw=black] (c0_stack3) at ([shift={(-2pt,2pt)}]c0_stack4) {};
    \node[claim_box, fill=white, draw=black] (c0_stack2) at ([shift={(-2pt,2pt)}]c0_stack3) {};
    \node[claim_box, fill=white, draw=black] (c0_stack1) at ([shift={(-2pt,2pt)}]c0_stack2) {};
    \node[claim_box, fill=white, draw=black] (c0_main)   at ([shift={(-2pt,2pt)}]c0_stack1) {Claim $c_n$};
    \node[stack_container, fit=(c0_stack3) (c0_main)] (stack0_box) {};

    \node[claim_box, fill=rot!20, draw=rot!50!black] (c1_stack4) [below=2.5cm of stack0_box, xshift=4pt] {};
    \node[claim_box, fill=grun!20, draw=grun!50!black] (c1_stack3) at ([shift={(-2pt,2pt)}]c1_stack4) {};
    \node[claim_box, fill=rot!20, draw=rot!50!black] (c1_stack2) at ([shift={(-2pt,2pt)}]c1_stack3) {};
    \node[claim_box, fill=gray!30, draw=gray!70!black] (c1_stack1) at ([shift={(-2pt,2pt)}]c1_stack2) {};
    \node[claim_box, fill=grun!20, draw=grun!50!black] (c1_main) at ([shift={(-2pt,2pt)}]c1_stack1) {Claim $c_n$};
    \node[stack_container, fit=(c1_stack3) (c1_main)] (stack1_box) {};

    \node[claim_box, fill=grun!20, draw=grun!50!black] (c2_stack2) [below=1.35cm of stack1_box, xshift=2pt, yshift=-3pt] {};
    \node[claim_box, fill=gray!30, draw=gray!70!black] (c2_stack1) at ([shift={(-2pt,2pt)}]c2_stack2) {};
    \node[claim_box, fill=grun!20, draw=grun!50!black] (c2_main)   at ([shift={(-2pt,2pt)}]c2_stack1) {Claim $c_n$};
    \node[stack_container, fit=(c2_stack2) (c2_main)] (stack2_box) {};

    \node[response_style, below=1.05cm of stack2_box] (final) {\hspace{25pt}Rewritten response $R^*$};
    \node[inner_icon, draw=petrol, text=petrol, anchor=west, xshift=6pt] at (final.west) {!};

    \node[draw, rectangle, minimum width=6.6cm, minimum height=3cm,
          inner xsep=0.2cm, inner ysep=6pt, line width=0.8pt, fill=white,
          rounded corners=5pt, right=0.7cm of stack0_box, yshift=-2.8cm] (detail_box) {
        \begin{minipage}[t]{6.2cm}
            {\centering
            \begin{tikzpicture}[>=Latex, font=\sffamily\small]
                \node[claim_box, rectangle, fill=white, draw=black, rounded corners=0] (ci) {Claim $c_n$};

                \node[rectangle, minimum width=0.5cm, minimum height=1.1cm, text centered, line width=0.8pt, below=0.7cm of ci] (placeholder_q) {...};
                \node[claim_box, rectangle, fill=rwth-10, draw=rwth, rounded corners=0, left=0.2cm of placeholder_q, minimum width=2.4cm, text=rwth] (q1) {Question $q_{n,1}$};
                \node[claim_box, rectangle, fill=rwth-10, draw=rwth, rounded corners=0, right=0.2cm of placeholder_q, minimum width=2.4cm, text=rwth] (qm) {Question $q_{n,J}$};

                \node[rectangle, minimum width=0.5cm, minimum height=1.1cm, text centered, line width=0.8pt, below=0.5cm of placeholder_q] (placeholder_a) {...};
                \node[claim_box, rectangle, fill=petrol-10, draw=petrol, rounded corners=0, left=0.2cm of placeholder_a, minimum width=2.4cm, text=petrol] (a1) {Answer $y_{n,1,1}$};
                \node[claim_box, rectangle, fill=petrol-10, draw=petrol, rounded corners=0, right=0.2cm of placeholder_a, minimum width=2.4cm, text=petrol] (ar) {Answer $y_{n,1,K}$};

                \node[rectangle, minimum width=0.5cm, minimum height=0.5cm, text centered, below=0.5cm of placeholder_a] (placeholder_l) {...};
                \node[claim_box, rectangle, fill=grun!20, draw=grun!50!black, rounded corners=0, left=0.2cm of placeholder_l, minimum width=2.4cm] (l1) {entails};
                \node[claim_box, rectangle, fill=rot!20, draw=rot!50!black, rounded corners=0, right=0.2cm of placeholder_l, minimum width=2.4cm] (lr) {contradicts};

                \draw[->] (ci.south) -- (q1.north);
                \draw[->] (ci.south) -- (qm.north)
                    node[midway, right, overlay, font=\small, xshift=-2cm] {generate};
                \draw[->] (q1.south) -- (a1.north);
                \draw[->] (q1.south) -- (ar.north)
                    node[midway, right, overlay, font=\small, xshift=-1.3cm]
                    {sample $M_\text{base}$};
                \draw[->] (a1.south) -- (l1.north);
                \draw[->] (ar.south) -- (lr.north);
            \end{tikzpicture}\par}
            
            \noindent\makebox[\linewidth][c]{\begin{minipage}[t]{5.9cm}
                \vspace{0.05cm}
                \small 
                \sffamily  
                \noindent
                A question is \emph{entailing} (\emph{contradicting}) if at least $k_e$ ($k_c$) of its answers entail (contradict) $c_n$.\\
                $c_n$ is \textcolor{grun!50!black}{known} iff at least $j_e$ questions are entailing and at most $j_c$ are contradicting (\Cref{eq:recall_criterion}); otherwise it is \textcolor{rot}{unknown}.
                \vspace{0.1cm}
            \end{minipage}}
        \end{minipage}
    };
    \node[font=\itshape\small, below=2pt of detail_box.south]
        {\textcolor{petrol}{\textsf{\bfseries\upshape D}}\enspace Knowledge Probing (expanded)};

    \draw[->, line width=0.6pt] (stack0_box.north |- flame.south) --
        node[left=6pt, font=\itshape\small, midway] {decomposition}
        (stack0_box.north);

    \path (stack1_box.north |- stack0_box.south) -- (stack1_box.north) coordinate[midway] (zoom_pos);
    \draw[->, line width=0.6pt] (stack1_box.north |- stack0_box.south) -- (stack1_box.north)
        node[right=6pt, font=\itshape\small, midway,
             align=left, text width=1.55cm]
        {knowledge-\\dependent:\\probe};

    \coordinate (bypass_corner) at ([xshift=-1cm]stack0_box.west);
    \draw[->, line width=0.6pt, draw=gray!70!black, rounded corners=3pt]
        (stack0_box.west) -- (bypass_corner) --
        node[midway, right=3pt, font=\itshape\small,
             text=gray!70!black, rotate=-90, anchor=south]
        {non-knowledge-dependent}
        (bypass_corner |- stack2_box.west) -- (stack2_box.west);
    
    \node[circle, draw=petrol, text=petrol, line width=1pt, minimum size=15pt,
          inner sep=0pt, fill=petrol-10, font=\sffamily\scriptsize\bfseries]
          (zoom_circle) at ([xshift=0pt]zoom_pos) {D};

    \draw[->, line width=0.6pt] (stack2_box.north |- stack1_box.south) --
        node[right=6pt, font=\itshape\small, midway, align=left]
        {filtering out\\\textcolor{rot}{unknown}}
        (stack2_box.north);

    \draw[->, line width=0.6pt] (stack2_box.south) --
        node[left=6pt, font=\itshape\small, align=right, text width=2cm]
        {summarizing\\with $M_\text{rewriter}$}
        (stack2_box.south |- final.north);

    \end{tikzpicture}
    }\hspace*{0.05cm}}
\fi
    \caption{Recall Rewrite. Each knowledge-dependent claim of the gold response $R$ is probed with generated questions, answers sampled from $M_\text{base}$, and an entailment check of each answer against the claim. Claims classified as unknown are filtered out. The remaining content is rewritten as $R^*$.}
    \label{fig:confidence-rewrite-illustration}
\end{figure}

\emph{Recall Rewrite}, our primary contribution (\Cref{fig:confidence-rewrite-illustration}), approximates $\mathcal{K}(M_{\text{base}})$ without external evidence: it probes whether the base model can consistently recall each claim and treats such claims as known.
In contrast to UNIT, which relies on token-level confidence scores, we estimate knowledge through probing questions.
Our method adapts QA-based factual consistency ideas \cite{honovich-etal-2021-q2, manakul-etal-2023-selfcheckgpt} to closed-book, claim-level probing of the base model's parametric knowledge.

Given a prompt $P$ and its gold response $R$, we partition $\mathcal{C}(R \mid P)$ into \textit{knowledge-dependent} claims, which encode verifiable factual, procedural, or structural information (including meta-knowledge), and \textit{non-knowledge-dependent} claims, which are purely contextual, subjective, or rely on general reasoning.
The latter require no parametric knowledge and are therefore always considered known and preserved.

For each knowledge-dependent claim $c_n$, an auxiliary teacher model generates $J$ diverse, context-independent probing questions $\{q_{n,j}\}_{j=1}^J$ (\Cref{fig:claim_question_rewrite}).
For each question, we sample $K$ answers $\{y_{n,j,k}\}_{k=1}^K$ from $M_{\text{base}}$.
Only these answers come from $M_{\text{base}}$. Question generation, the entailment judgment below, and the final rewriting are performed by the teacher model (\Cref{tab:notation}; instantiated in \Cref{subsec:experimental_settings}).

Then, for each question--answer pair we evaluate whether $y_{n,j,k}$ entails, contradicts, or is unrelated to the original claim $c_n$ (\Cref{fig:template_entailment_check}).
Let $e_{n,j}$ and $d_{n,j}$ denote how many of the $K$ answers to question $q_{n,j}$ entail, respectively contradict, $c_n$.
A question inherits the label that enough of its answers carry: we call it \emph{entailing} if it yields at least $k_e$ entailing answers ($e_{n,j} \ge k_e$) and \emph{contradicting} if it yields at least $k_c$ contradicting answers ($d_{n,j} \ge k_c$).
A knowledge-dependent claim $c_n$ is \emph{consistently recalled} by $M_\text{base}$ iff
\begin{equation}
\label{eq:recall_criterion}
\begin{aligned}
\text{(i)}\;\;  & \underbrace{\bigl|\{\,j : e_{n,j} \ge k_e\,\}\bigr|}_{\text{entailing questions}} \ge j_e
\quad\text{and} \\[2pt]
\text{(ii)}\;\; & \underbrace{\bigl|\{\,j : d_{n,j} \ge k_c\,\}\bigr|}_{\text{contradicting questions}} \le j_c .
\end{aligned}
\end{equation}
Condition (i) requires the claim to be recovered from several independently phrased questions, guarding against probes that leak the answer.
Condition (ii) tolerates a small number of contradictions, which typically stem from underspecified probes (\Cref{sec:cr_filter_appendix}).
Recall Rewrite classifies $c_n$ as \emph{known} iff it is consistently recalled and as \emph{unknown} otherwise.

Finally, a rewriter model $M_{\text{rewriter}}$ produces the rewritten response $R^*$ by removing any content that entails or implies a claim classified as \textit{unknown}.
The rewriter preserves the structure, style, and non-knowledge-dependent claims of the original response.
As in Evidence Rewrite, $M_\text{rewriter}$ may return a refusal if the remaining information is insufficient.

\begin{table*}[!t]
\centering
\small
\begin{tabular}{l c rrrr rrrr}
\toprule
& & \multicolumn{4}{c}{WildHalu} & \multicolumn{4}{c}{Bios} \\
\cmidrule(lr){3-6} \cmidrule(lr){7-10}
Approach & Dataset & \#Ref. & \#Supp. & \%Supp. & FActScore & \#Ref. & \#Supp. & \%Supp. & FActScore \\
\midrule
\qweninstruct{} & - & 85 & {8,722} & {85.9} & {87.1} & 243 & {6,100} & {60.5} & {79.3} \\
Standard SFT & T\"ulu 3 & 1 & 6,452 & 79.1 & 77.2 & 1 & 2,689 & 34.4 & 33.3 \\
\midrule
Standard SFT & OASST1 & 2 & \underline{8,059} & 76.6 & 74.4$^{***}$ & 4 & \underline{5,061} & 36.0 & 34.1$^{***}$ \\
FLAME & OASST1 & 6 & 7,882 & 73.0 & 74.4$^{***}$ & 6 & \textbf{5,066} & 34.0 & 33.4$^{***}$ \\
UNIT$_{cut}$ & OASST1 & 0 & \textbf{8,111} & \underline{81.2} & \underline{79.4}$^{*}$ & 3 & 5,043 & \underline{45.1} & \underline{43.1}$^{***}$ \\

Evidence Rewrite (ours) & OASST1 & 1 & 7,842 & 80.1 & 78.3 & 6 & 3,663 & 42.3 & 39.9 \\

\emph{- w/o brainstorming} & OASST1 & 4 & 6,344 & 79.2 & 76.8 & 14 & 2,453 & 37.8 & 36.7 \\

Recall Rewrite (ours) & OASST1 & 55 & 6,157 & \textbf{84.2} & \textbf{84.1} & 252 & 2,561 & \textbf{56.2} & \textbf{76.4} \\
\bottomrule
\end{tabular}
\caption{Factuality results for all SFT approaches using \qwenmodel{}. The top block shows the fully post-trained instruct model from Alibaba\notehere{\qweninstructid} and a large-data SFT baseline (T\"ulu~3); the bottom block compares knowledge-aligned methods on OASST1. \#Ref.: number of refusals; \#Supp.: number of claims in non-refusal responses judged supported; \%Supp.: percentage of claims in non-refusal responses that are supported; FActScore: mean per-example supported-claim ratio, treating refusals as fully supported. Significance markers ($^{***}$~$p < 0.001$, $^{*}$~$p < 0.05$) indicate Wilcoxon signed-rank tests against Recall Rewrite\notehere{differences between Recall Rewrite and \qweninstruct{} were not statistically significant, $p > 0.05$, on either dataset}. Best results of models on OASST1 are \textbf{bolded}; second best are \underline{underlined}.}
\label{tab:main_results}
\end{table*}

\section{Experiments}

\subsection{Experimental Settings}
\label{subsec:experimental_settings}

\paragraph{Training Details.} We utilize \qwenbaseid{} \cite{qwen3} as the base model $M_{\text{base}}$. %
Following \citet{flame}, we use the English subset of \emph{Open Assistant 1} (OASST1) \cite{oasst} as the SFT dataset, which consists of crowd-sourced multi-turn conversation trees.
We follow the standard approach of using the first turn in each conversation tree, resulting in 3,468 data points.
All models are trained using SFT implemented via the TRL library \cite{trl}, maintaining consistent hyperparameters across all variants (see \Cref{sec:tab:train_params}).
Specifically, we train standard SFT, FLAME, UNIT$_{cut}$, Evidence Rewrite, and Recall Rewrite models.
Additional baseline implementation details are provided in \Cref{sec:baseline_method_details}.

\setcounter{footnote}{1}
\notetext{\qweninstructid}
\setcounter{footnote}{2}
\notetext{Differences between Recall Rewrite and \qweninstruct{} were not statistically significant ($p > 0.05$) on either dataset.}

\paragraph{Evidence Rewrite Model.}
We apply the Evidence Rewrite pipeline from \Cref{subsec:factuality_rewrite} to OASST1 using \gptfouromini{} \citepalias{gpt4o} for all pipeline steps.
The fact-checking pipeline comprises three components: \emph{VeriScore claim decomposition} \cite{song-etal-2024-veriscore}, a hierarchical \emph{evidence retrieval} step (\Cref{subsec:fact-checking}), and \emph{FActScore claim verification} \cite{min-etal-2023-factscore}.
We use VeriScore rather than FActScore decomposition because FActScore frequently produces unverifiable claims on open-domain instruction-following data, such as claims with unresolved pronouns or meta-commentary about the response itself, consistent with observations by \citet{song-etal-2024-veriscore}.
Evidence is retrieved from Wikipedia.
The final rewriting step is performed by \gptfouromini{} (\Cref{fig:factuality-rewrite-illustration}).
 
\paragraph{Recall Rewrite Model.}
We apply the Recall Rewrite pipeline from \Cref{subsec:confidence_rewrite} to OASST1 using \gptfivemini{} for all stages (claim decomposition, question generation, entailment check and response rewriting).
For each knowledge-dependent claim, we generate five probing questions ($J{=}5$) and sample two responses per question ($K{=}2$) from $M_{\text{base}}$ at temperature $0.5$ using a few-shot QA prompt (\Cref{fig:inference}).
As the default filter in \Cref{eq:recall_criterion} we use $j_e{=}2, k_e{=}1, j_c{=}2, k_c{=}1$, abbreviated as $j_e/k_e/j_c/k_c = 2/1/2/1$.
A claim is thus known iff at least two questions receive an entailing answer and at most two questions receive any contradicting answer.
Statistics for the application of the pipeline are reported in \Cref{tab:cr-stats}.
Under the default filter, 79.4\% (\qwenbaseid{}) of the knowledge-dependent claims are classified as known and retained, so that together with the non-knowledge-dependent claims 87--88\% of all claims survive the rewrite.
API costs are listed in \Cref{tab:cr_api_cost}.
We release the resulting knowledge-aligned training data for both base models together with all intermediate pipeline outputs, i.e. the decomposed claims, probing questions, base-model answers and entailment labels\footnote{\label{fn:data-release}\url{https://huggingface.co/datasets/apptek-com/recall-rewrite-oasst1}}.

\paragraph{Factuality Evaluation.}
We follow the evaluation framework of \citet{wu2025balancingtruthfulnessinformativenessuncertaintyaware}, which combines two long-form factual generation benchmarks: WildHalu \cite{wildhalu} and Biography generation (Bios) \cite{min-etal-2023-factscore}.
WildHalu evaluates responses about 500 real-world entities, about half of which lack Wikipedia pages, spanning domains such as computing, finance, culture, and geography.
Verification uses evidence retrieved from Google Search.
Biography asks models to answer \emph{\enquote{Question: Tell me a bio of X}} for 500 people with Wikipedia pages, using the linked Wikipedia page as the sole evidence source in the retrieval phase.

Responses are fact-checked by decomposing them into atomic claims, retrieving evidence, and verifying entailment as described in detail in \Cref{subsec:fact-checking}.
Following \citet{wu2025balancingtruthfulnessinformativenessuncertaintyaware}, we use \gptfour{} for claim decomposition and \gptfouromini{} for claim verification.
We report \emph{\#Supp.}, the number of supported claims in non-refusal responses; \emph{\%Supp.}, the percentage of claims in non-refusal responses that are supported; and \emph{FActScore}, the per-example percentage of supported claims with refusals treated as fully supported.
Claims the verifier labels as \texttt{Not known} are excluded from all three metrics (\Cref{subsec:fact-checking}).

\begin{table*}[h]
\centering
\small
\begin{tabular}{l l rrrr rrrr}
\toprule
& & \multicolumn{4}{c}{WildHalu} & \multicolumn{4}{c}{Bios} \\
\cmidrule(lr){3-6} \cmidrule(lr){7-10}
Model & Data & \#Ref. & \#Supp. & \%Supp. & FActScore & \#Ref. & \#Supp. & \%Supp. & FActScore \\
\midrule
SFT & Dolci & 7 & 6,939 & 77.9 & 75.6 & 9 & 4,638 & 30.7 & 34.7 \\
DPO & Dolci & 25 & \underline{7,566} & 76.6 & 76.5 & 200 & \underline{5,230} & 45.7 & \underline{66.3} \\
RLVR & Dolci & 29 & 7,235 & \underline{78.5} & \underline{78.4} & 229 & 4,860 & \textbf{51.8} & \textbf{72.4} \\
\midrule
SFT (ours) & OASST1 & 6 & \textbf{7,830} & 77.9 & 75.8 & 3 & \textbf{5,347} & 37.3 & 35.2 \\
RR (ours) & OASST1 & 40 & 6,730 & \textbf{82.7} & \textbf{82.5} & 120 & 3,730 & \underline{47.3} & 56.6 \\
\bottomrule
\end{tabular}
\caption{Results on WildHalu and Biography for \olmoinstruct{} across post-training stages compared to our variants trained on OASST1. The top block shows official checkpoints trained on Dolci; the bottom block shows our reproduced SFT baseline and Recall Rewrite (RR). Best results per column are \textbf{bolded}; second best are \underline{underlined}.}
\label{tab:olmo_results}
\end{table*}

\begin{table*}
\centering
\small
\begin{tabular}{rr rrrr rrrr}
\toprule
\multicolumn{2}{c}{Train (OASST1)} & \multicolumn{4}{c}{WildHalu} & \multicolumn{4}{c}{Bios} \\
\cmidrule(lr){1-2} \cmidrule(lr){3-6} \cmidrule(lr){7-10}
\%Known & Avg.Cl. & \#Ref. & \#Supp. & \%Supp. & FActScore & \#Ref. & \#Supp. & \%Supp. & FActScore \\
\midrule
100 & 16.5 & 97 & 5,418 & \textbf{84.7} & \textbf{86.1} & 214 & 2,679 & \textbf{51.8} & \textbf{69.9} \\
50 & 13.8 & 45 & \textbf{5,806} & 78.7 & 80.4 & 13 & \textbf{4,268} & 39.5 & 38.7 \\
0  & 12.6 & 22 & 4,458 & 79.0 & 79.5 & 4 & 3,680 & 39.1 & 38.4 \\
\bottomrule
\end{tabular}
\caption{Effect of the share of known claims in the Recall Rewrite training targets on factuality (\qwenmodel{}). \%Known is the percentage of known claims among the retained knowledge-dependent claims (not the filter share of \Cref{tab:cr_filters}); the number of non-refusal (1,777) and refusal (302) training examples is identical in all rows. Avg.Cl.\ reports claims per response in training. Evaluation columns as in \Cref{tab:main_results}. Best results per column are \textbf{bold}.}
\label{tab:cr_ablations}
\end{table*}

\subsection{Main Results}
\label{subsec:main_results}

\Cref{tab:main_results} summarizes the performance of all methods on the WildHalu and Biography tasks. Across both datasets, all knowledge-aligned variants (except FLAME) consistently outperform standard SFT in terms of \%Supp. and FActScore, indicating reduced hallucinations. This is consistent with the premise that aligning supervision with the model's parametric knowledge reduces hallucinations, and \Cref{subsec:cr_ablation} tests it directly.

Recall Rewrite achieves the highest \%Supp. and FActScore on both datasets, but at the cost of fewer supported claims (\#Supp.) and substantially increased refusal rates.
In contrast, UNIT$_{cut}$ and Evidence Rewrite retain more supported claims but achieve lower \%Supp., reflecting a consistent trade-off between hallucination reduction and coverage: stricter filtering yields more reliable but less informative outputs.
Refusal behavior is central to this trade-off because FActScore treats refusals as fully supported.
Methods with higher FActScore, particularly Recall Rewrite and \qweninstruct{}, exhibit higher refusal rates, especially on the Biography task, which contains many long-tail facts.
A per-entity comparison (\Cref{subsec:qual_generations}) shows that these refusals almost exclusively concern entities on which standard SFT hallucinates, while over-refusals on well-known entities do occur on WildHalu.
Recall Rewrite's higher FActScore should therefore not be interpreted as improved factual generation at equal coverage.
It instead reflects a more conservative response policy: the fraction of supported claims among non-refusal responses rises, while the total number of supported claims falls.

Comparing generation- and estimation-based approaches, FLAME does not improve over standard SFT, suggesting that naive generation is an unreliable proxy for parametric knowledge. Adding filtering or verification (Evidence Rewrite) consistently improves over FLAME. Similarly, while UNIT$_{cut}$ outperforms generation-based methods, Recall Rewrite achieves higher measured factuality, suggesting a more reliable but more conservative knowledge-alignment strategy.

The official \qweninstruct{} model remains stronger than all OASST1-trained variants.
This comparison is not controlled: the official model differs in instruction data scale and composition, post-training stages, and possible benchmark contamination.
Our goal is therefore not to outperform the complete post-training recipe of \qweninstruct{}, but to isolate whether knowledge-aligned SFT data improves factuality under a shared training setup.
The \tulusft{} \cite{lambert2025tulu} baseline suggests that simply increasing SFT data scale does not remove the problem.

\subsection{Multi-Stage Post-Training}

For \qweninstruct{} the data mixture and intermediate checkpoints after SFT are not available.
To better understand the potential impact of different training stages, we turn to \olmoinstruct{} \citepalias{olmo2026olmo3}, for which public checkpoints after SFT, preference optimization (DPO) and reinforcement learning with verifiable rewards (RLVR) are available.
\Cref{tab:olmo_results} shows the results for different post-training checkpoints of the 7B parameter model.
Compared to SFT, the DPO checkpoint increases the number of supported claims (\#Supp.) while also increasing refusals.
The RLVR checkpoint further improves \%Supp. and FActScore.
These trends suggest that additional training stages can improve factual reliability.

For comparison, we train a standard SFT and Recall Rewrite model on OASST1 starting from the \olmomodel{} base model.
Although Recall Rewrite relies solely on SFT with knowledge-aligned data, it consistently achieves higher \%Supp. and FActScore on WildHalu than all OLMo checkpoints, while maintaining a comparable number of supported claims.
Taken together, these results suggest that, in our setup, knowledge-aligned data construction can achieve gains comparable to additional post-training stages; whether these gains stack with factuality-oriented RL remains open.

\begin{table*}
\centering
\small
\begin{tabular}{l cccc cccc cccc}
    \toprule
    & \multicolumn{4}{c}{{FalseQA}} & \multicolumn{4}{c}{{NEC}} & \multicolumn{4}{c}{{RefuNQ}} \\
    \cmidrule(lr){2-5} \cmidrule(lr){6-9} \cmidrule(lr){10-13}
    {Model} & {A} & {P} & {R} & {F} & {A} & {P} & {R} & {F} & {A} & {P} & {R} & {F} \\
    \midrule

    Standard SFT & 
    72.6 & 83.1 & 56.6 & 67.4 & 
    66.1 & \textbf{79.0} & 43.9 & 56.4 & 
    63.3 & 67.7 & 50.7 & 58.0 \\

    FLAME & 
    71.8 & 82.4 & 55.5 & 66.4 & 
    65.6 & 77.7 & 43.9 & 56.1 & 
    63.9 & 67.8 & 53.1 & 59.5 \\

    UNIT$_{cut}$ & 
    71.1 & 77.8 & 59.1 & 67.2 & 
    61.7 & 72.5 & 37.7 & 49.7 & 
    64.8 & 65.9 & 61.7 & 63.7 \\
    
    Evidence Rewrite (ours) & 
    72.0 & \textbf{83.5} & 54.8 & 66.2 & 
    63.5 & 76.4 & 39.1 & 51.8 & 
    62.6 & 67.2 & 49.3 & 56.9 \\

    \emph{- w/o brainstorming} & 
    \textbf{73.0} & 82.8 & 58.1 & 68.3 & 
    67.0 & 75.0 & 50.9 & 60.6 & 
    65.8 & \textbf{68.1} & 59.6 & 63.6 \\

    Recall Rewrite (ours) & 
    70.8 & 74.0 & \textbf{64.1} & \textbf{68.7} & 
    \textbf{69.9} & 71.4 & \textbf{66.3} & \textbf{68.8} & 
    \textbf{65.9} & 62.6 & \textbf{79.1} & \textbf{69.9} \\

    \bottomrule
\end{tabular}
\caption{Results on \emph{UnknownBench}. All scores are percentages. A = Accuracy, P = Precision, R = Recall, F = F1 score. P is the share of refusals that occur on unanswerable prompts, and R is the share of unanswerable prompts that are refused. The highest value in each column is highlighted in \textbf{bold}.}
\label{tab:refusal_benchmark}
\end{table*}

\begin{table*}
\centering
\small
\begin{tabular}{l l rrrrr}
\toprule
Approach & Dataset & Avg. & HumanEval+ & GSM8K & IFEval & TruthfulQA \\
\midrule
\qweninstruct{} & - & {80.7} & {87.8} & {90.8} & {86.0} & {58.2} \\
Standard SFT & \tulusft{} & {74.2} & {89.3} & 80.7 & {71.7} & 54.9 \\
\midrule
Standard SFT & OASST1 & \textbf{69.8} & 83.8 & \textbf{83.6} & \textbf{57.5} & \underline{54.4}   \\
FLAME & OASST1 & \underline{69.4} & \textbf{87.8} & 81.3 & 54.3 & 54.2 \\
UNIT$_{cut}$ & OASST1 & 68.5 & 83.0 & 81.6 & \underline{56.6} & 52.6 \\
Evidence Rewrite (ours) & OASST1 & 68.2 & 85.6 & \underline{82.6} & 52.3 & 52.2 \\
\emph{- w/o brainstorming} & OASST1 & 67.7 & \underline{85.9} & 78.2 & 53.4 & 53.1  \\
Recall Rewrite (ours) & OASST1 & 68.9 & 84.3 & 81.3  & 54.3 & \textbf{55.5} \\
\bottomrule
\end{tabular}
\caption{Comparison of training approaches on general capability benchmarks. All methods are based on \qwenmodel{} unless otherwise specified. Best results on OASST1 are \textbf{bolded}; second best are \underline{underlined}.}
\label{tab:results_olmes}
\end{table*}

\subsection{Varying the Share of Known Claims}
\label{subsec:cr_ablation}

To test whether more knowledge-aligned training targets indeed improve factuality, we vary only the proportion of knowledge-dependent claims classified as \emph{known} in the targets (\%Known), keeping the prompts, the number of non-refusal (1,777) and refusal (302) examples, and all training settings fixed.
Non-knowledge-dependent claims are always retained (\Cref{subsec:confidence_rewrite}).
Among the knowledge-dependent claims of each response, we retain the largest randomly sampled subset that matches the target ratio of known to unknown claims.
At 100\%, only known knowledge-dependent claims are retained, at 50\% equally many known and unknown claims, and at 0\% only unknown claims.

\Cref{tab:cr_ablations} shows that increasing \%Known improves \%Supp. and FActScore, with the 100\% setting achieving the strongest reduction in hallucinations.
However, this comes at the cost of fewer supported claims (\#Supp.) and increased refusal rates.
The 50\% setting retains more supported claims but yields lower \%Supp. and FActScore.
These trends mirror the main results in \Cref{tab:main_results} and highlight a consistent trade-off: stricter knowledge alignment reduces hallucinations but limits informativeness. 
\Cref{fig:tradeoff_coverage} (\Cref{sec:cr_filter_appendix}) visualizes this coverage--factuality trade-off for all models, including the filter-threshold ablation.
The one factor we do not control is the number of retained claims per response (Avg.Cl.), which is lower at 0\% because unknown claims are rarer than known ones.

\subsection{Refusal Behavior}

Evaluating refusal behavior provides a complementary perspective on hallucinations: a model should ideally refuse unanswerable questions while responding to those within its knowledge boundary.
To evaluate this dimension, we use \emph{UnknownBench} \cite{unknownbench}, which comprises three subtasks.
Each subtask pairs answerable with unanswerable prompts, and the subtasks differ in how the unanswerable questions are constructed.
\emph{FalseQA} consists of 4,730 questions where the unanswerable instances assume non-existent relations between entities, i.e., they are based on false premises.
\emph{NEC} (Non-Existing Concepts) contains 4,144 instances and challenges the model to identify plausible but fabricated concepts based on invented words.
The unanswerable prompts for \emph{RefuNQ} (4,346 instances total) are created by replacing random nouns with non-existent concepts.
To ensure a balanced evaluation, we randomly downsampled instances from each subtask so that the classes of answerable and unanswerable questions are equal in size.

\Cref{tab:refusal_benchmark} presents the results on UnknownBench.
Recall Rewrite, which refuses most frequently in the factuality evaluation (\Cref{subsec:main_results}), achieves substantially higher recall across all subtasks, indicating that it refuses more reliably when a question is unanswerable.
Conversely, Recall Rewrite exhibits the lowest precision, reflecting a higher rate of false refusals. While these results suggest a trade-off between precision and recall, Recall Rewrite outperforms the alternative methods on all subtasks in terms of F1-Score. \Cref{fig:tradeoff_unknownbench} (\Cref{sec:unknownbench_pr_appendix}) plots the precision--recall operating points of all models together with the \%Known ablation. %
Qualitative examples of correct refusals and over-refusals, both in the training data and in the model generations, are given in \Cref{sec:qualitative_appendix}.

\subsection{Performance on General Tasks}
\label{subsec:general_tasks}

Beyond hallucination-focused benchmarks, we evaluate general model capabilities using four tasks implemented via \emph{OLMES} \cite{olmes}.
\emph{HumanEval+} \cite{evalplus} is a coding benchmark built around manually crafted test cases. We compare models using the \verb|pass@10| metric.
\emph{GSM8K} \cite{gsm8k} assesses mathematical reasoning across 8.5K grade school math problems, measuring exact matches between gold labels and model generations.
\emph{IFEval} \cite{ifeval} comprises 541 prompts with verifiable instructions and evaluates instruction-following capability via prompt-level accuracy.
Finally, \emph{TruthfulQA} \cite{lin-etal-2022-truthfulqa} tests models on 817 multiple-choice questions regarding common human misconceptions. We report the percentage of questions where the model identifies \emph{all} correct answers (MC2).

The results are presented in \Cref{tab:results_olmes}.
Our methods perform comparably to standard SFT, indicating that the factuality gains do not come with clear degradation on the general capability benchmarks considered here.
\Cref{fig:tradeoff_general} (\Cref{sec:tradeoff_general_appendix}) summarizes this trade-off by plotting FActScore (\Cref{tab:main_results}) against the OLMES average.
Recall Rewrite improves FActScore over standard SFT by 10 points on WildHalu and 42 points on Bios.
Its OLMES average is only 0.9 points lower, within the 2.1-point band spanned by all OASST1 models.
The largest per-benchmark differences of Recall Rewrite to standard SFT (GSM8K $-2.3$, IFEval $-3.2$) are of the same magnitude as those among the baselines themselves (e.g., FLAME $-3.2$ on IFEval, Evidence Rewrite w/o brainstorming $-5.4$ on GSM8K).
A per-example analysis (\Cref{subsec:qual_olmes}) shows that the GSM8K errors are ordinary reasoning slips without any refusals, which points to training variance, whereas the IFEval difference is fully explained by Recall Rewrite refusing 30 of the 541 prompts, mostly creative-writing tasks.
On the remaining IFEval prompts both models are on par (56.3 vs.\ 56.5).
As expected, the model trained on the much larger T\"ulu~3 mixture and the official \qweninstruct{} model clearly outperform the models trained on the smaller OASST1 dataset.

\section{Related Work}

\paragraph{Factual hallucination detection.}
Hallucinations are commonly divided into inconsistencies with the input context and contradictions of real-world facts \cite{survey-hallucinations, survey-hallucinations-nlg}.
We focus on the latter.
Detecting factual hallucinations typically requires decomposing generations into atomic claims, retrieving evidence, and verifying claim support, as in FActScore \cite{min-etal-2023-factscore} and later open-domain variants using web search \cite{safe, song-etal-2024-veriscore}.
When external evidence is unavailable, prior work has used model self-evaluation or confidence-based signals as proxies for factuality \cite{llm-know-what-they-know, facttune, self-consistency, fadeeva-etal-2024-fact}.
Concurrent to our work, \citet{calderon2026emptyshelves} treat a fact as known to a model only if it consistently answers differently phrased questions about it, mirroring our consistent-recall criterion (\Cref{subsec:confidence_rewrite}), and find that recall, rather than encoding, is the main bottleneck for parametric factuality.

\paragraph{Hallucinations from fine-tuning.}
Controlled studies show that fine-tuning on facts unknown to the base model increases hallucinations: at the example level in closed-book QA \cite{gekhman-etal-2024-fine}, for long-form generation \cite{flame}, and at the claim level in instruction data \cite{wu2025balancingtruthfulnessinformativenessuncertaintyaware}.
\citet{flame} and \citet{wu2025balancingtruthfulnessinformativenessuncertaintyaware} also propose mitigations, replacing gold targets with base-model generations and filtering claims by uncertainty, respectively.
We frame these mitigations as approximations to knowledge-aligned SFT and compare them under a unified setup.
\Cref{subsec:cr_ablation} reproduces the effect of \citet{gekhman-etal-2024-fine} in open-domain instruction data.

\paragraph{Alternative mitigation strategies.}
Other approaches reduce hallucinations through mechanisms orthogonal to knowledge-aligned SFT.
\citet{kaplan2026finetuningencourageshallucinationsfix} attribute SFT-induced hallucinations to \emph{factual forgetting}, i.e., interference with pre-existing factual representations, and counteract it with optimization-level regularization and reduced factual plasticity.
Instead of filtering SFT supervision to match existing model knowledge, \citet{liu2025fictitious} inject relevant facts through continued pre-training before SFT.
\citet{thulke-etal-2025-listen} show that removing training examples that are unfaithful to the retrieved context improves faithfulness in retrieval-augmented generation, applying an analogous target-curation strategy with respect to the provided context rather than parametric knowledge.
\citet{tan2026selfimprovingpretrainingusingposttrained} also use a strong post-trained model to rewrite training data, but apply rewriting during pre-training and not SFT.
Selective refusal methods \cite{huang-etal-2025-alleviating} and preference-based objectives such as DPO \cite{flame, facttune, huang-chen-2024-factalign, kto} provide complementary ways to improve factuality.

\section{Conclusion}

We framed hallucinations from SFT as a consequence of mismatch between the factual claims required by training targets and the parametric knowledge of the base model.
When targets require unsupported knowledge, SFT may teach the model to answer confidently beyond its knowledge boundary.
Knowledge-aligned SFT addresses this by aligning training targets with the model's knowledge boundary.
Our comparison shows that how this boundary is approximated matters.
Naive self-generation is not sufficient: FLAME does not improve factuality over standard SFT in our setting.
External verification improves generation-based alignment, but the largest metric gains in our experiments come from Recall Rewrite, which directly probes whether the base model can consistently recall the claims used for supervision.

Across \qwenmodel{} and \olmomodel{}, and both the broader WildHalu setting and the Wikipedia-centered Biography evaluation, knowledge-aligned supervision reduces measured hallucinations without clear losses on the general capability benchmarks considered here.
Refusal behavior on UnknownBench improves at the cost of lower factual coverage and more false refusals.
Additional comparisons suggest that the gains are not explained solely by SFT data size, and that knowledge-aligned data construction remains competitive with stronger post-training stages.
These results do not imply that knowledge-aligned SFT replaces full instruct-model post-training recipes, but rather that it may target a distinct source of hallucinations that such recipes do not always address.

Overall, knowledge-aligned rewriting makes the factual content of SFT targets a controllable variable, and controlling it reduces hallucination behavior.
Applying this control to large instruction-tuning pipelines will require cheaper approximations of the knowledge probing than our current pipeline provides.

\section*{Limitations}

Our framework treats a model's parametric knowledge as an object that can be approximated from model behavior, but it cannot be observed directly.
Both generation-based and recall-based methods therefore provide approximations of $\mathcal{K}(M_{\text{base}})$.
In addition, our formulation largely treats factual knowledge as a binary property: a claim is either classified as known (and retained) or unknown (and removed).
Real models may instead exhibit graded confidence, partial knowledge, or sensitivity to phrasing.
Modeling these distinctions more explicitly could further improve knowledge-aligned data construction.
Relatedly, our method comparisons in \Cref{tab:main_results} are correlational with respect to the composition of the training targets.
Only the ablation in \Cref{subsec:cr_ablation} manipulates the share of known claims while holding the training set size and refusal ratio fixed, and it does so for a single base model and dataset.

Our evaluation also depends on automatic factuality measurement.
It is furthermore entity-centric: following standard practice in long-form factuality evaluation, WildHalu and Biography both elicit descriptions of a single named entity.
WildHalu extends beyond Wikipedia-covered entities (about half of its entities lack Wikipedia pages \cite{wildhalu}), but neither benchmark tests prompts whose factual content is not organized around one entity.
The claim decomposition, evidence retrieval, and verification stages can introduce errors, especially for claims that are underspecified, difficult to retrieve, or require domain expertise.
This limitation is shared with much recent work on long-form factuality evaluation.
To reduce its impact, we use established fact-checking pipelines and apply the same evaluation procedure across systems, so the main comparisons are based on a consistent measurement protocol.

We focus on supervised fine-tuning and do not fully characterize how knowledge-aligned SFT interacts with later post-training stages.
The OLMo comparison suggests that factuality-oriented DPO or RLVR can improve hallucination behavior, but we do not test whether the gains from Recall Rewrite stack with such objectives.
Studying these combinations is an important direction for future work, especially for systems that already include strong factuality-oriented preference or reinforcement learning stages.

Scalability is another limitation.
Recall Rewrite requires claim decomposition, probing question generation, repeated base-model sampling, entailment checking, and response rewriting, making it more useful as a high-precision diagnostic intervention than as a complete scalable recipe for all SFT data preparation.
The question-generation step also becomes more difficult for complex responses that combine many facts, reasoning steps, or implicit background assumptions.
Our experiments use the English first-turn subset of OASST1, which is small compared to contemporary instruction-tuning mixtures.
As a result, our findings should be tested on larger instruction-tuning mixtures and in settings such as multilingual or multi-turn dialogue, tool use, coding, reasoning-heavy prompts, and specialized domains.

Finally, our rewriting pipelines rely on strong teacher models.
These models can introduce their own biases in claim selection, question generation, verification, refusal style, and the phrasing of rewritten answers.
Using multiple teachers, open-weight teachers, or human audits could help separate teacher-specific effects from the effects of aligning supervision with the base model's parametric knowledge.

\makeatletter%
\ifacl@anonymize\else%
\section*{Acknowledgments}
Generative AI assistants (OpenAI Codex and Claude Code) were used to proofread the manuscript, to help with LaTeX formatting, and to support with the implementation of the experiments.
All AI-generated suggestions were reviewed and verified by the authors, who take full responsibility for the content of this work.

\fi%
\makeatother%

\bibliography{custom, anthology_part1, anthology_part2, anthology_part3, fact_rewrite}

\newpage
\appendix

\section*{Appendix Overview}

The appendix is organized as follows.
\Cref{sec:glossary} collects definitions of the closely related terms used throughout the paper, and \Cref{sec:unified_algorithm} states the data-construction algorithm shared by all knowledge-aligned SFT methods.
\Cref{sec:baseline_method_details,subsec:fact-checking,sec:tab:train_params} describe the baseline methods, the fact-checking pipeline, and the training setup, including statistics and cost of the Recall Rewrite pipeline.
\Cref{sec:cr_filter_appendix,sec:tradeoff_general_appendix,sec:unknownbench_pr_appendix,sec:qualitative_appendix} present additional results: the filter threshold ablation, the factuality--capability trade-off, the UnknownBench precision--recall trade-off, and a qualitative analysis of Recall Rewrite.
Finally, \Cref{sec:prompts_appendix} lists all prompts used in this work.

\section{Glossary}
\label{sec:glossary}

The paper uses several closely related terms.
This glossary collects their definitions and where they are introduced.

\begin{description}
\item[Known / unknown claim] Framework-level property of a claim $c$: known iff $c \in \mathcal{K}(M_\text{base})$ (\Cref{sec:framework}).
Since $\mathcal{K}(M_\text{base})$ is latent, each method approximates this property, and we write \enquote{classified as known/unknown} for the outcome.
\item[Consistently recalled] Recall Rewrite's operational test for a claim (\Cref{eq:recall_criterion}): at least $j_e$ probing questions are entailing and at most $j_c$ are contradicting, where a question inherits a label if enough of its answers from the base model carry it.
Recall Rewrite classifies a claim as known iff it is consistently recalled.
\item[Knowledge-dependent claim] A claim that encodes verifiable factual, procedural, or structural information and therefore requires parametric knowledge (\Cref{subsec:confidence_rewrite}).
Non-knowledge-dependent claims (contextual, subjective, or generic) are always retained, and only knowledge-dependent claims are probed.
\item[Knowledge-seeking prompt] A prompt that cannot be answered without factual knowledge (\Cref{subsec:existing_work}; details in \Cref{sec:baseline_method_details}).
FLAME and Evidence Rewrite keep the gold response for prompts that are not knowledge-seeking.
\item[Supported / unsupported claim] Verdict of fact-checking a claim against retrieved external evidence (\Cref{subsec:fact-checking}), used to build the Evidence Rewrite training data and, at evaluation time, to compute \#Supp., \%Supp.\ and FActScore.
This is independent of known/unknown: a claim can be supported by evidence yet unknown to the base model, and vice versa.
\end{description}

\section{Unified Algorithm Framework}
\label{sec:unified_algorithm}

\Cref{tab:notation} summarizes the models by the role they play and the outputs they produce.
\Cref{alg:unified} states the data-construction procedure shared by all knowledge-aligned SFT methods considered in this paper, covering claim decomposition, the detection or retrieval step, filtering, rewriting, and the refusal decision, and \Cref{tab:unified_instantiations} lists how each method instantiates its three hooks.

\begin{table}[h]
\centering
\small
\setlength{\tabcolsep}{4pt}
\begin{tabular}{@{}lll@{}}
\toprule
Symbol & Role & Output space \\
\midrule
$M_\text{base}$ & base model & text \\
$M_\text{SFT}$  & fine-tuned on $\mathcal{D}^*$ & text \\
\midrule
$M_\text{KS}$      & knowledge-seeking   & $\{\textbf{true}, \textbf{false}\}$ \\
                    & prompt classifier   &  \\
$M_\text{dec}$      & claim decomposition & $\{c_1,\dots,c_N\}$ \\
$M_\text{verifier}$ & claim vs.\ evidence & \texttt{\{SUPPORTED,} \\
                    &                     & \texttt{NOT\_SUPPORTED\}} \\
$M_\text{probe}$    & probe generation    & $\{q_{n,1},\dots,q_{n,J}\}$ \\
$M_\text{judge}$    & answer vs.\ claim   & \texttt{\{ENTAILS,} \\
                    &                     & \texttt{CONTRADICTS,} \\
                    &                     & \texttt{UNRELATED\}} \\
$M_\text{rewriter}$ & composes the target & text \\
\bottomrule
\end{tabular}
\caption{Models by role in \Cref{alg:unified}. The first block contains the models under study; the second block contains pipeline components used only to construct $\mathcal{D}^*$, which are never trained. Evidence Rewrite instantiates $M_\text{KS}$, $M_\text{dec}$, $M_\text{verifier}$ and $M_\text{rewriter}$ with \gptfouromini{}; Recall Rewrite instantiates $M_\text{dec}$, $M_\text{probe}$, $M_\text{judge}$ and $M_\text{rewriter}$ with \gptfivemini{} and uses no $M_\text{KS}$ (\Cref{subsec:experimental_settings}).}
\label{tab:notation}
\end{table}

\begin{algorithm*}[!tp]
\small
\caption{Knowledge-aligned SFT data construction. Both proposed pipelines and both baselines instantiate one template, differing in the hooks \textsc{Gate}, \textsc{Source} and \textsc{Unknown} and in the models $M_\text{dec}$ and $M_\text{rewriter}$ (\Cref{tab:unified_instantiations,tab:notation}).}
\label{alg:unified}
\begin{algorithmic}[1]
\Require $\mathcal{D}=\{(P,R)\}$
    \Comment{SFT dataset of prompt--response pairs}
\Statex \hspace{\algorithmicindent}$M_\text{base}$
    \Comment{base model whose parametric knowledge $\mathcal{K}(M_\text{base})$ is the alignment target}
\Statex \hspace{\algorithmicindent}$M_\text{dec}$
    \Comment{decomposition model, realizes $\mathcal{C}(\,\cdot \mid P)$}
\Statex \hspace{\algorithmicindent}$M_\text{rewriter}$ (optional)
    \Comment{if set, composes the aligned target $R^*$, or a refusal}
\Statex \hspace{\algorithmicindent}\textsc{Gate}$(P)$
    \Comment{hook: is the prompt processed at all?}
\Statex \hspace{\algorithmicindent}\textsc{Source}$(P,R)$
    \Comment{hook: which response is decomposed}
\Statex \hspace{\algorithmicindent}\textsc{Unknown}$(c_n)$
    \Comment{hook: flags a claim for removal}
\Ensure knowledge-aligned data $\mathcal{D}^*$
\State $\mathcal{D}^* \gets \emptyset$
\ForAll{$(P,R) \in \mathcal{D}$}
    \If{\textbf{not} \textsc{Gate}$(P)$}
        \State $\mathcal{D}^* \gets \mathcal{D}^* \cup \{(P,R)\}$; \textbf{continue}
            \Comment{gold response kept unchanged}
    \EndIf
    \Statex
    \State $S \gets \textsc{Source}(P,R)$
        \Comment{gold $R$, or $\hat{R} \sim M_\text{base}$}
    \Statex
    \If{$M_\text{rewriter}$ is not set}
        \State $\mathcal{D}^* \gets \mathcal{D}^* \cup \{(P,S)\}$; \textbf{continue}
            \Comment{source response used unchanged (FLAME)}
    \EndIf
    \Statex
    \State $\{c_1,\dots,c_N\} \gets M_\text{dec}(S \mid P)$
        \Comment{claim decomposition $\mathcal{C}(S \mid P)$}
    \State $\mathcal{C}_{\text{keep}} \gets \{\, c_n : \neg\,\textsc{Unknown}(c_n) \,\}$
        \Comment{keep all but flagged claims}
    \Statex
    \State $R^* \gets M_\text{rewriter}(P, S, \mathcal{C}_{\text{keep}})$
        \Comment{fluent response over $\mathcal{C}_{\text{keep}}$, or a refusal}
    \Statex
    \State $\mathcal{D}^* \gets \mathcal{D}^* \cup \{(P,R^*)\}$
\EndFor
\State \Return $\mathcal{D}^*$

\Statex
\Procedure{Unknown}{$c_n$} \textbf{-- Evidence Rewrite}\label{lin:unknown_er}
    \State $E_n \gets \textsc{Retrieve}(c_n)$
        \Comment{top-5 Wikipedia chunks}
    \State \Return $\bigl[\, M_\text{verifier}(c_n, E_n) = \texttt{NOT\_SUPPORTED} \,\bigr]$
\EndProcedure

\Statex
\Procedure{Unknown}{$c_n$} \textbf{-- Recall Rewrite}\label{lin:unknown_rr}
    \If{\textbf{not} \textsc{KnowledgeDependent}$(c_n)$}
        \State \Return \textbf{false}
            \Comment{labeled by $M_\text{dec}$; no parametric knowledge needed}
    \EndIf
    \State $\{q_{n,j}\}_{j=1}^{J} \gets M_\text{probe}(c_n)$
        \Comment{context-independent probing questions}
    \State $y_{n,j,k} \sim M_\text{base}(q_{n,j})$ \textbf{for} $j \in [J],\, k \in [K]$
    \State $e_{n,j} \gets \bigl|\{k : M_\text{judge}(y_{n,j,k}, c_n) = \texttt{ENTAILS}\}\bigr|$
    \State $d_{n,j} \gets \bigl|\{k : M_\text{judge}(y_{n,j,k}, c_n) = \texttt{CONTRADICTS}\}\bigr|$
    \State $\textit{consistentlyRecalled} \gets \bigl[\,|\{j : e_{n,j} \ge k_e\}| \ge j_e \,\wedge\, |\{j : d_{n,j} \ge k_c\}| \le j_c\,\bigr]$
        \Comment{\Cref{eq:recall_criterion}}
    \State \Return $\neg\,\textit{consistentlyRecalled}$
\EndProcedure
\end{algorithmic}
\end{algorithm*}

\begin{table*}
\centering
\footnotesize
\setlength{\tabcolsep}{2.5pt}
\renewcommand{\cellalign}{tl}
\begin{tabular}{@{}llllll@{}}
\toprule
Method & \textsc{Gate}$(P)$ & \textsc{Source}$(P,R)$ & $M_\text{dec}$ & \textsc{Unknown}$(c_n)$ & $M_\text{rewriter}$ \\
\midrule\midrule
FLAME & $M_\text{KS}(P)$ & $\hat{R} \sim M_\text{base}(P)$ & --- & always \textbf{false} & --- \\
\midrule
UNIT$_{cut}$ & always \textbf{true} & gold $R$ & \makecell{\citeauthor{wu2025balancingtruthfulnessinformativenessuncertaintyaware}\\\citeyearpar{wu2025balancingtruthfulnessinformativenessuncertaintyaware}} & $\mathrm{CCP}(c_n) \le \tau$ & \makecell{\citeauthor{wu2025balancingtruthfulnessinformativenessuncertaintyaware}\\\citeyearpar{wu2025balancingtruthfulnessinformativenessuncertaintyaware}} \\
\midrule
\makecell{Evidence Rewrite\\(ours)} & \makecell{$M_\text{KS}(P)$} & \makecell{$\hat{R} \sim M_\text{base}(P)$, then\\$\hat{R} \sim M_\text{base}(\textsc{BrainstormPrompt}(P,\hat{R}))$} & \makecell{\Cref{fig:claim_decomposition_prompt}} & \makecell{Alg.~\ref{alg:unified}, l.~\ref{lin:unknown_er}} & \makecell{\Cref{fig:summarization_prompt_template}} \\
\midrule
\makecell{Recall Rewrite\\(ours)} & always \textbf{true} & gold $R$ & \Cref{fig:confidence_rewrite_decomp} & Alg.~\ref{alg:unified}, l.~\ref{lin:unknown_rr} & \Cref{fig:template_redaction} \\
\bottomrule
\end{tabular}
\caption{Instantiations of \Cref{alg:unified}: the methods differ in the three hooks, in $M_\text{dec}$ and in $M_\text{rewriter}$. For our two methods $M_\text{dec}$ and $M_\text{rewriter}$ are instantiated by the prompt templates referenced in the corresponding cells, as is \textsc{BrainstormPrompt} (\Cref{fig:brainstorm_example}); \textsc{Gate} is implemented by the knowledge-seeking classifier $M_\text{KS}$ (\Cref{fig:factual_prompt_classifier}), which skips the prompt when it returns \textbf{false}; FLAME sets no rewriter (---) and uses its sampled response unchanged.}
\label{tab:unified_instantiations}
\end{table*}

\section{Baseline Method Details}
\label{sec:baseline_method_details}

\paragraph{FLAME.} Previous work \cite{gekhman-etal-2024-fine, flame} provides empirical evidence that training on unknown information can lead to a model learning to hallucinate more. \citet{flame} propose \emph{FLAME} as a mitigation method whose core idea is to use responses sampled from the base model instead of the gold response $R$ for knowledge-seeking prompts. The motivation is that $R$ may contain information outside the base model's parametric knowledge, whereas a sampled response $\hat{R} \sim M_{\text{base}}$ is more likely to reflect what the base model can already generate.

In instruction fine-tuning datasets such as OASST1, some prompts do not require the model to draw from its internal knowledge (e.g., summarization and paraphrasing). Therefore, for these prompts, FLAME retains the gold response $R$. In our replication, we identify such prompts using a classifier based on \gptfouromini{} \citepalias{gpt4o} with few-shot prompting (\Cref{fig:factual_prompt_classifier}).

For knowledge-seeking prompts, \emph{FLAME} samples $\hat{R}$ from $M_{\text{base}}$ in a few-shot setting. The few-shot setting is necessary because the base model has not been trained to follow instructions.

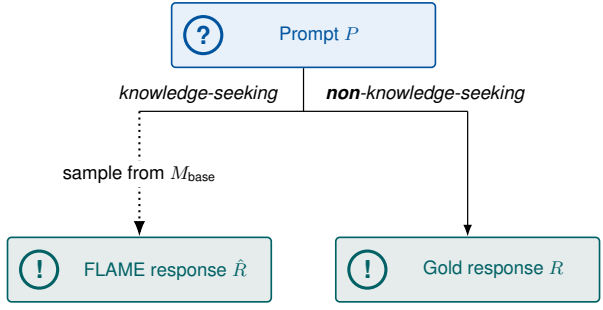
\begin{figure}
    \centering
    \resizebox{0.5\textwidth}{!}{
\iflatexml
\htmlfig{flame}{Flow diagram of FLAME. Knowledge-seeking prompts are replaced by a response sampled from the base model; non-knowledge-seeking prompts keep the gold response.}%
\else
    \begin{tikzpicture}[
        >=Latex, 
        font=\sffamily\small,
        node distance=1.5cm and 0cm,
        base/.style={
            draw, rectangle, 
            minimum width=4.5cm, minimum height=1.1cm, 
            text centered, line width=0.8pt, rounded corners=2pt
        },
        prompt_style/.style={base, fill=rwth-10, draw=rwth, text=rwth},
        response_style/.style={base, fill=petrol-10, draw=petrol, text=petrol},
        claim_box/.style={
            draw, rectangle, 
            minimum width=1.5cm, minimum height=0.5cm, 
            text centered, font=\sffamily\small, line width=0.5pt
        },
        stack_container/.style={draw, dashed, inner sep=10pt, rounded corners=3pt, color=gray!50},
        inner_icon/.style={
            circle, draw, inner sep=0pt, 
            font=\sffamily\large\bfseries, 
            line width=1.5pt, 
            minimum size=18pt
        }
    ]

    \node[prompt_style] (prompt) at (0, 5.5) {\hspace{15pt}Prompt $P$};
    \node[inner_icon, draw=rwth, text=rwth, anchor=west, xshift=6pt] at (prompt.west) {?};

    \coordinate (branch_center) at (0, 4.2);
    \draw[line width=0.6pt] (prompt.south) -- (branch_center);
    
    \draw[line width=0.6pt] (-2.8, 4.2) -- (2.8, 4.2);

    \node[response_style] (flame) at (-2.8, 1.5) {\hspace{25pt}FLAME response $\hat{R}$};
    \node[inner_icon, draw=petrol, text=petrol, anchor=west, xshift=6pt] at (flame.west) {!};
    
    \draw[->, dotted, line width=1.0pt] (-2.8, 4.2) -- 
        node[left=0.5cm, pos=0, anchor=south west, font=\sffamily\small\itshape, align=left] {knowledge-seeking}
        node[fill=white, inner sep=2pt, pos=0.5] {sample from $M_\text{base}$} 
        (flame.north);

    \node[response_style, at={(2.8, 1.5)}] (gold) {\hspace{25pt}Gold response $R$};
    \node[inner_icon, draw=petrol, text=petrol, anchor=west, xshift=6pt] at (gold.west) {!};
    
    \draw[->, line width=0.6pt] (2.8, 4.2) -- 
        node[right=1.1cm, pos=0, anchor=south east, font=\sffamily\footnotesize\itshape, align=left] {\textbf{non}-knowledge-seeking}
        (gold.north);

    \end{tikzpicture}
    }
\fi
    \caption{FLAME. For knowledge-seeking prompts, the response $\hat{R}$ is sampled from the base model $M_\text{base}$.}
    \label{fig:flame-illustration}
\end{figure}

When preparing FLAME training data for OASST1, we retrieve the top five similar but not identical training examples for each prompt $P$ using \verb|bge-large-en-v1.5| \cite{bge}. We remove less relevant examples to fit into $\frac{3}{4}$ of the context size; if fewer than three examples remain, the prompt is removed from the training set.
\Cref{fig:flame-illustration} illustrates this data construction procedure.

\paragraph{UNIT$_{cut}$}
\citet{wu2025balancingtruthfulnessinformativenessuncertaintyaware} investigate the trade-off between informativeness and truthfulness in instruction fine-tuning, attributing hallucinations to the inclusion of unfamiliar, long-tail knowledge in training data. To address this, they propose uncertainty-aware instruction fine-tuning, where uncertainty estimates are used as a proxy for whether a model knows a given piece of information. In \emph{UNIT$_{cut}$}, each gold response $R$ is decomposed into atomic claims $\mathcal{C}(R \mid P)$, and claims with high estimated uncertainty are removed before training to obtain a filtered response $R^*$. This filtering improves truthfulness but often reduces informativeness due to the removal of content.

However, this approach relies on uncertainty as an imperfect proxy for model knowledge, which may be miscalibrated or only weakly correlated with parametric knowledge. Furthermore, UNIT$_{cut}$ operates solely on explicitly stated claims and does not account for implicit knowledge required to construct a valid response. In particular, responses often depend on \emph{meta-knowledge}: factual, procedural, or structural information that is not directly expressed in $R$ but is nonetheless required for generation. These limitations motivate a more direct estimation of model knowledge that captures both explicit and implicit knowledge.

For data preparation, we use the code provided by the authors\footnote{\url{https://github.com/AndrewWTY/UNIT}} and train the models using the same training pipeline as all other models.

\section{Fact-Checking Pipeline}
\label{subsec:fact-checking}

The fact-checking pipeline is used in two places in this paper: (i) to construct the Evidence Rewrite training data by verifying claims in base model generations, and (ii) to evaluate all models on the WildHalu and Biography benchmarks. In both cases, the pipeline follows the three-step approach of \citet{min-etal-2023-factscore}.

\paragraph{Claim Decomposition.}
The generated text is first segmented into sentences using \texttt{nltk} \cite{bird-loper-2004-nltk}. A sliding window is then applied to extract atomic claims from each sentence via LLM inference. The primary objective is to produce a comprehensive list of verifiable atomic claims — individual statements that can be independently checked against an external source. Different implementations vary in how they define atomicity, leading to different prompt templates \cite{min-etal-2023-factscore, song-etal-2024-veriscore} or additional filtering steps \cite{safe}.

We use the VeriScore decomposition prompt \cite{song-etal-2024-veriscore} rather than the original FActScore prompt \cite{min-etal-2023-factscore}. FActScore's decomposition was designed for biography generation and performs poorly on open-domain instruction-following data such as OASST1: it frequently extracts claims that cannot be verified against any external source. Common failure cases include claims with unresolved pronouns (e.g., \emph{"He has appeared in numerous films."}) and meta-commentary about the response itself (e.g., \emph{"There are provided passages."} extracted from \emph{"Based on the provided passages, here is the answer to the question:"}). VeriScore addresses this by using detailed instructions and few-shot examples that guide the model to only extract claims that are fully self-contained and can be checked against an external source, consistent with observations by \citet{song-etal-2024-veriscore}.

\paragraph{Evidence Retrieval.}
For Evidence Rewrite training data construction and Biography evaluation, the top five pieces of supporting evidence are retrieved from Wikipedia using a two-phase hierarchical approach. First, the Wikimedia API\footnote{\url{https://api.wikimedia.org/}} identifies the five most relevant Wikipedia pages for the claim. Second, these pages are split into chunks and reranked using \texttt{gtr-t5-large} \cite{ni-etal-2022-large} to select the five most pertinent chunks as evidence. We use the Wikipedia dump from 04/01/2023 provided by \citet{min-etal-2023-factscore}, preprocessed into fixed-size chunks. For WildHalu evaluation, we instead follow \citet{wu2025balancingtruthfulnessinformativenessuncertaintyaware} and retrieve evidence from Google Search using their implementation\footnote{\url{https://github.com/AndrewWTY/UNIT}}.

\paragraph{Claim Verification.}
Each claim is paired with its retrieved evidence chunks and passed to an LLM-based classifier, which decides whether the claim is supported by the evidence.
We use \gptfouromini{} \citepalias{gpt4o} for this step in both cases, but the two use cases differ in the label set.
For Evidence Rewrite training data construction, we use the FActScore verification prompt (\Cref{fig:claim_verification_prompt}), which labels each claim as \texttt{SUPPORTED} or \texttt{NOT\_SUPPORTED}.
For evaluation, we follow the implementation of \citet{wu2025balancingtruthfulnessinformativenessuncertaintyaware}, whose verifier distinguishes three labels: \texttt{True}, \texttt{False}, and \texttt{Not known} for claims that the retrieved evidence neither confirms nor refutes\footnote{These three labels are also the ones reported for the qualitative examples in \Cref{sec:qualitative_appendix}.}.
We treat \texttt{True} as supported and \texttt{False} as unsupported, and exclude claims labeled \texttt{Not known} from \#Supp., \%Supp.\ and FActScore.

\section{Training Details}
\label{sec:tab:train_params}

\begin{table}[h]
\centering
\small
\begin{tabular}{lr}
\toprule
Parameter & Value \\
\midrule
Devices & 1 $\times$ NVIDIA A100 (80GB) \\
Number of Epochs & 5 \\
Total Batch Size & 32 \\
Learning Rate & $1 \times 10^{-5}$ \\
Optimizer & AdamW \\
Scheduler & Cosine \\
Warmup Ratio & 0.1 \\
Weight Decay & 0.1 \\
Context Length & 1024 \\
\bottomrule
\end{tabular}
\caption{Training configuration for SFT training on OASST1.}
\label{tab:train_params}
\end{table}

\Cref{tab:train_params} provides the SFT configuration used for standard SFT and the different knowledge-aligned SFT approaches.

\subsection{Recall Rewrite}

\begin{table}[h]
\centering
\small
\setlength{\tabcolsep}{3pt}
\begin{tabular}{lrr}
\toprule
Metric & \qwenmodel{} & \olmomodel{} \\
\midrule
Total examples & 3,467 & 3,467 \\
Total claims & 64,301 & 64,301 \\
Avg. claims per example & 18.5 & 18.5 \\
\midrule
Entailed answers & 62.5\% & 70.6\% \\
Unrelated answers & 25.1\% & 15.3\% \\
Contradicting answers & 12.4\% & 14.1\% \\
\midrule
Non-knowl.-dep. claims & 24,448 (38.0\%) & 24,448 (38.0\%) \\
Knowl.-dep. claims & 39,853 (62.0\%) & 39,853 (62.0\%) \\
\quad known (retained) & 31,654 (79.4\%) & 32,168 (80.7\%) \\
\quad unknown (removed) & 8,199 (20.6\%) & 7,685 (19.3\%) \\
Retained claims & 56,102 (87.2\%) & 56,616 (88.0\%) \\
Avg. claims after rewrite & 16.5 & 16.6 \\
\midrule
Non-refusal responses & 3,165 & 3,175 \\
Refusal responses & 302 & 292 \\
\bottomrule
\end{tabular}
\caption{Statistics of the Recall Rewrite pipeline on OASST1 for \qwenbaseid{} and \olmobaseid{} under the default filter $2/1/2/1$ (\Cref{eq:recall_criterion})\notehere{one of the 3,468 examples is dropped by the pipeline because its claim decomposition could not be parsed, an invalid JSON escape, leaving 3,467}. Answer shares are over all probe answers. Percentages for non-/knowledge-dependent and retained claims are relative to all claims, those for known/unknown claims relative to the knowledge-dependent claims. Non-knowledge-dependent claims are always retained, so the retained claims are the non-knowledge-dependent plus the known claims. Avg.\ claims after rewrite counts retained claims per non-refusal response.}
\label{tab:cr-stats}
\end{table}
\notetext{One of the 3,468 examples is dropped by the pipeline because its claim decomposition could not be parsed (invalid JSON escape), leaving 3,467.}

\begin{table*}
\centering
\small
\begin{tabular}{lrrr}
\toprule
Pipeline Step & \# Input Tokens & \# Output Tokens & Cost (USD) \\
\midrule
Claim Decomposition & 3 M & 1.2 M & 1.98 \\
Claim-Question Rewriting & 63.5 M & 19.7 M & 27.64\\
Entailment Check & 82.5 M & 0.6 M & 10.92 \\
Answer Rewriting & 23.5 M & 0.5 M & 3.44 \\
\midrule
\addlinespace
\textbf{Total} & 172.5 M & 22 M & 43.98 \\
\bottomrule
\end{tabular}
\caption{Token usage and API cost of the Recall Rewrite pipeline per step, using
\gptfiveminifull{} with batched API calls across all 3,468 OASST1 examples
for \qwenmodel{}.}
\label{tab:cr_api_cost}
\end{table*}

\Cref{tab:cr-stats} provides statistics on applying the Recall Rewrite pipeline using both \qwenmodel{} and \olmomodel{} on the OASST1 data.
Of the 39,853 knowledge-dependent claims, 8,199 (\qwenbaseid{}) and 7,685 (\olmobaseid{}) are classified as unknown and removed, while all 24,448 non-knowledge-dependent claims are retained.
Refusals concentrate on responses with a low share of known claims: among the knowledge-dependent claims of refusal responses only 60.9\% (\qwenbaseid{}) and 59.8\% (\olmobaseid{}) are known, compared to 81.7\% and 83.2\% in non-refusal responses.
Accordingly, non-refusal responses shrink from 18.5 to 16.5 (\qwenbaseid{}) and 16.6 (\olmobaseid{}) claims on average.
\Cref{tab:cr_api_cost} provides the corresponding token usage and API cost for the different steps.

\section{Recall Rewrite: Filter Threshold Ablation}
\label{sec:cr_filter_appendix}

\begin{table*}
\centering
\small
\setlength{\tabcolsep}{2.5pt}
\begin{tabular}{rrrrr rrrr rrrr}
\toprule
\multicolumn{5}{c}{Train} & \multicolumn{4}{c}{WildHalu} & \multicolumn{4}{c}{Bios} \\
\cmidrule(lr){1-5} \cmidrule(lr){6-9} \cmidrule(lr){10-13}
Filter & \%Known & \#Non-Ref. & Avg.Cl. & \#Ref. & \#Ref. & \#Supp. & \%Supp. & FActScore & \#Ref. & \#Supp. & \%Supp. & FActScore \\
\midrule
2/1/2/1 & 79.4 & 3,165 & 16.5 & 100 & 32 & 6,006 & 82.8 & 82.1 & 66 & 2,989 & 43.9 & 49.7 \\
\midrule
2/1/5/3 & 84.7 & 3,273 & 16.9 & 100 & 15 & \textbf{6,877} & 80.5 & 78.4 & 38 & \textbf{3,870} & 43.2 & 42.9 \\
2/1/0/1 & 62.3 & 2,877 & 14.9 & 100 & 25 & 5,312 & \textbf{84.3} & \textbf{82.6} & 40 & 2,802 & 49.3 & 49.3 \\
\midrule
1/1/2/1 & 83.2 & 3,211 & 16.9 & 100 & 11 & 6,603 & 80.9 & 78.9 & 34 & 3,854 & 43.0 & 45.3 \\
3/2/2/1 & 53.6 & 2,781 & 13.9 & 100 & 30 & 5,206 & 82.3 & 81.1 & 65 & 3,198 & \textbf{49.8} & \textbf{53.2} \\
\bottomrule
\end{tabular}
\caption{Effect of the Recall Rewrite filter thresholds $j_e/k_e/j_c/k_c$ (\Cref{eq:recall_criterion}: at least $j_e$ entailing and at most $j_c$ contradicting questions) on \qwenbaseid{} trained on OASST1. The first row is the default used in the main paper. \%Known is the share of knowledge-dependent claims in the training targets that the filter classifies as known and retains (cf.\ \Cref{tab:cr-stats}; not the target ratio of \Cref{tab:cr_ablations}); Avg.Cl.\ is the number of claims per non-refusal training response after rewriting. The number of refusals in the training data (\#Ref.\ Train) is held fixed at 100. Best results are \textbf{bolded}.}
\label{tab:cr_filters}
\end{table*}

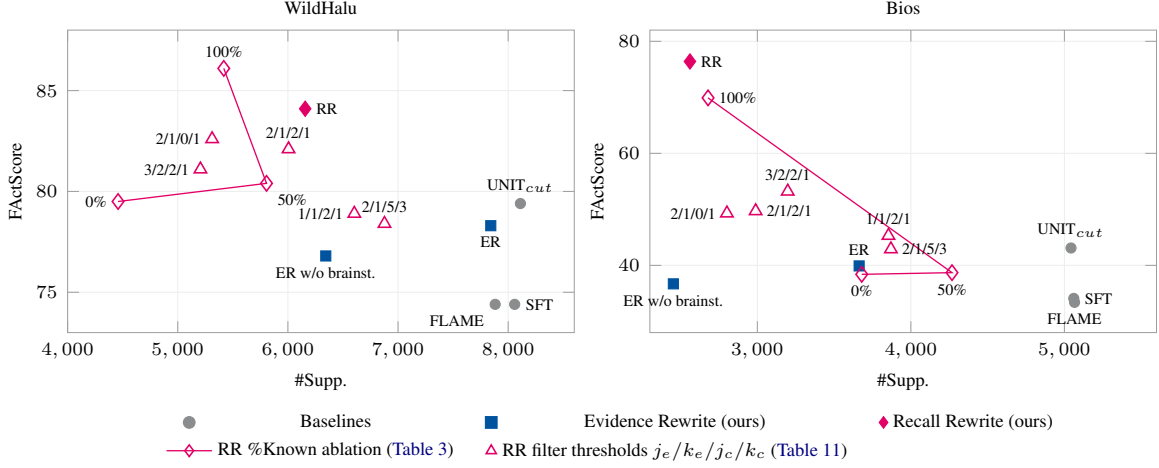
\begin{figure*}[t]
\centering
\iflatexml
\htmlfig{tradeoff_coverage}{Two scatter plots of coverage (number of supported claims) against FActScore, for WildHalu and Bios. Recall Rewrite and its stricter ablations sit up and to the left (higher factuality, lower coverage); baselines sit to the right.}%
\else
\begin{tikzpicture}
\begin{groupplot}[
  tradeoff, width=6.7cm, height=4.0cm,
  group style={group size=2 by 1, horizontal sep=1.0cm},
  ylabel={FActScore}, xlabel={\#Supp.},
  scaled x ticks=false, xticklabel style={/pgf/number format/1000 sep={,}, /pgf/number format/fixed},
  tolegend,
]
\nextgroupplot[legend to name=tradeoffCoverageLegend, legend columns=3, title={WildHalu}, xmin=4000, xmax=8600, ymin=73, ymax=88, xtick distance=1000]
\addplot[pt/base, tolabel] table[x=x, y=y, meta=label] {
x y label anchor
8059 74.4 SFT west
7882 74.4 FLAME {north east}
8111 79.4 \unitcut{} south
};
\addlegendentry{Baselines}
\addplot[pt/er, tolabel] table[x=x, y=y, meta=label] {
x y label anchor
7842 78.3 ER north
6344 76.8 {ER w/o brainst.} north
};
\addlegendentry{Evidence Rewrite (ours)}
\addplot[pt/rr, tolabel] table[x=x, y=y, meta=label] {
x y label anchor
6157 84.1 RR west
};
\addlegendentry{Recall Rewrite (ours)}
\addplot[pt/rrabl, tolabel] table[x=x, y=y, meta=label] {
x y label anchor
4458 79.5 0\% east
5806 80.4 50\% {north west}
5418 86.1 100\% south
};
\addlegendentry{RR \%Known ablation (\Cref{tab:cr_ablations})}
\addplot[pt/rrfilt, tolabel] table[x=x, y=y, meta=label] {
x y label anchor
6006 82.1 2/1/2/1 south
6877 78.4 2/1/5/3 south
5312 82.6 2/1/0/1 east
6603 78.9 1/1/2/1 east
5206 81.1 3/2/2/1 east
};
\addlegendentry{RR filter thresholds $j_e/k_e/j_c/k_c$ (\Cref{tab:cr_filters})}

\nextgroupplot[title={Bios}, xmin=2300, xmax=5600, ymin=28, ymax=82, xtick distance=1000]
\addplot[pt/base, tolabel] table[x=x, y=y, meta=label] {
x y label anchor
5061 34.1 SFT west
5066 33.4 FLAME north
5043 43.1 \unitcut{} south
};
\addplot[pt/er, tolabel] table[x=x, y=y, meta=label] {
x y label anchor
3663 39.9 ER south
2453 36.7 {ER w/o brainst.} north
};
\addplot[pt/rr, tolabel] table[x=x, y=y, meta=label] {
x y label anchor
2561 76.4 RR west
};
\addplot[pt/rrabl, tolabel] table[x=x, y=y, meta=label] {
x y label anchor
3680 38.4 0\% north
4268 38.7 50\% north
2679 69.9 100\% west
};
\addplot[pt/rrfilt, tolabel] table[x=x, y=y, meta=label] {
x y label anchor
2989 49.7 2/1/2/1 west
3870 42.9 2/1/5/3 west
2802 49.3 2/1/0/1 east
3854 45.3 1/1/2/1 south
3198 53.2 3/2/2/1 south
};
\end{groupplot}
\node[anchor=north] at ($(group c1r1.south west)!0.5!(group c2r1.south east) + (0,-0.8cm)$) {\pgfplotslegendfromname{tradeoffCoverageLegend}};
\end{tikzpicture}
\fi
\caption{Coverage vs.\ factuality on WildHalu (left) and Bios (right) for all \qwenmodel{} models trained on OASST1. \#Supp.\ is the number of supported claims over all non-refusal responses (coverage); FActScore treats refusals as fully supported. The two Recall Rewrite ablations trace the trade-off controlled at training time: the proportion of known claims in the targets (\%Known, hollow diamonds, \Cref{tab:cr_ablations}) and the filter thresholds $j_e/k_e/j_c/k_c$ (triangles, \Cref{tab:cr_filters}; trained with 100 instead of 302 refusal examples, hence not identical to the main Recall Rewrite model). Up and to the right is better.}
\label{fig:tradeoff_coverage}
\end{figure*}

We study how different $j_e/k_e/j_c/k_c$ filtering thresholds (defined in \Cref{subsec:confidence_rewrite}) affect the Recall Rewrite pipeline.
The default used in the main paper is $2/1/2/1$.
To isolate the effect of claim filtering, the number of refusals in the training data is held fixed at 100 across all conditions.
Results are shown in \Cref{tab:cr_filters}.

Across all conditions we observe the same trade-off as in the main ablation (\Cref{tab:cr_ablations}): more lenient thresholds yield more supported claims (\#Supp.) at evaluation but lower \%Supp.\ and FActScore, while stricter thresholds reduce \#Supp.\ but improve \%Supp.\ and FActScore.
\Cref{fig:tradeoff_coverage} plots both ablations together with all methods of \Cref{tab:main_results} in the coverage--factuality plane.
The Recall Rewrite variants span a wide range of operating points that no single baseline covers: the stricter settings ($3/2/2/1$, $2/1/0/1$, 100\% known) reach FActScore levels no baseline attains, whereas the two most lenient filter settings ($1/1/2/1$, $2/1/5/3$) are dominated by UNIT$_{cut}$, which retains more supported claims at a similar or higher FActScore.
The filter threshold thus acts as a training-time knob for choosing a point on this trade-off.

\paragraph{Varying the contradiction tolerance ($j_c/k_c$).}
One motivation for tolerating some contradicting answers is that probing questions are sometimes \emph{underspecified}: an overly open question may admit multiple valid answers that, when compared against a specific claim, are inconsistently labeled as \textit{Contradicts}.
Under zero contradiction tolerance ($2/1/0/1$), a single such label disqualifies the claim.
Compared to the default $2/1/2/1$, this results in fewer \#Supp.\ while \%Supp.\ and FActScore remain similar on WildHalu, and on Bios \%Supp.\ improves but FActScore does not --- suggesting that the additionally discarded claims are largely correct ones.
Going in the other direction, $2/1/5/3$ cannot be violated with $K{=}2$ answers per question and thus disables the contradiction criterion.
It yields more \#Supp.\ but lower \%Supp.\ and FActScore on both datasets, indicating that dropping the contradiction criterion retains genuinely unknown claims.

\paragraph{Varying the entailment requirement ($j_e/k_e$).}
One motivation for requiring entailment from multiple questions is that probing questions are sometimes \emph{overspecified}: a question that closely mirrors the wording of the original claim may elicit an entailing response even when the model lacks genuine knowledge.
Compared to the default $2/1/2/1$, relaxing to $1/1/2/1$ yields more \#Supp.\ but lower \%Supp.\ and FActScore, consistent with retaining claims the model does not robustly know.
Tightening to $3/2/2/1$ improves \%Supp.\ and FActScore, most notably on Bios, at the cost of discarding additional training claims.
At evaluation this reduces \#Supp.\ on WildHalu, whereas on Bios the stricter model retains slightly more supported claims than the default.

\paragraph{Choice of default ($2/1/2/1$).}
The default was motivated by two observed failure modes in question generation.
Requiring entailment from at least two distinct questions with at least one entailing response each ($j_e{=}2, k_e{=}1$) guards against overspecified probes.
Tolerating up to two questions with a single contradicting answer ($j_c{=}2, k_c{=}1$) guards against underspecified ones.
The results confirm that $2/1/2/1$ achieves a good balance: on WildHalu it retains more supported claims than the stricter settings while maintaining higher \%Supp.\ and FActScore than the more lenient ones.

\section{Factuality--Capability Trade-off}
\label{sec:tradeoff_general_appendix}

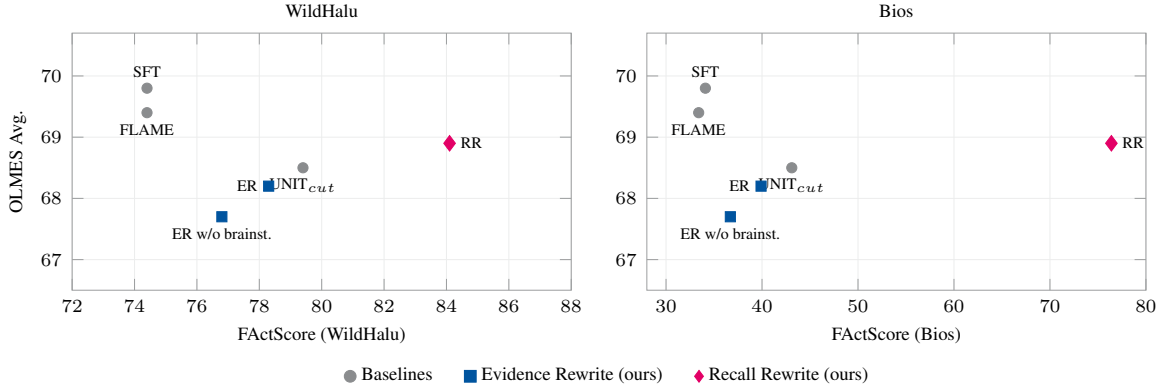
\begin{figure*}[t]
\centering
\iflatexml
\htmlfig{tradeoff_general}{Two scatter plots of FActScore against the OLMES average for WildHalu and Bios. All models lie within a 2.1-point band on the OLMES average while FActScore varies by 10 and 43 points.}%
\else
\begin{tikzpicture}
\begin{groupplot}[
  tradeoff, width=6.6cm, height=3.4cm,
  group style={group size=2 by 1, horizontal sep=1.0cm, ylabels at=edge left},
  ylabel={OLMES Avg.}, ymin=66.5, ymax=70.7,
  tolegend,
]
\nextgroupplot[legend to name=tradeoffGeneralLegend, title={WildHalu}, xlabel={FActScore (WildHalu)}, xmin=72, xmax=88]
\addplot[pt/base, tolabel] table[x=x, y=y, meta=label] {
x y label anchor
74.4 69.8 SFT south
74.4 69.4 FLAME north
79.4 68.5 \unitcut{} north
};
\addlegendentry{Baselines}
\addplot[pt/er, tolabel] table[x=x, y=y, meta=label] {
x y label anchor
78.3 68.2 ER east
76.8 67.7 {ER w/o brainst.} north
};
\addlegendentry{Evidence Rewrite (ours)}
\addplot[pt/rr, tolabel] table[x=x, y=y, meta=label] {
x y label anchor
84.1 68.9 RR west
};
\addlegendentry{Recall Rewrite (ours)}

\nextgroupplot[title={Bios}, xlabel={FActScore (Bios)}, xmin=28, xmax=80]
\addplot[pt/base, tolabel] table[x=x, y=y, meta=label] {
x y label anchor
34.1 69.8 SFT south
33.4 69.4 FLAME north
43.1 68.5 \unitcut{} north
};
\addplot[pt/er, tolabel] table[x=x, y=y, meta=label] {
x y label anchor
39.9 68.2 ER east
36.7 67.7 {ER w/o brainst.} north
};
\addplot[pt/rr, tolabel] table[x=x, y=y, meta=label] {
x y label anchor
76.4 68.9 RR west
};
\end{groupplot}
\node[anchor=north] at ($(group c1r1.south west)!0.5!(group c2r1.south east) + (0,-0.75cm)$) {\pgfplotslegendfromname{tradeoffGeneralLegend}};
\end{tikzpicture}
\fi
\caption{Factuality vs.\ general capability for all knowledge-alignment methods trained on OASST1 with \qwenmodel{}. Each point is one trained model: FActScore on WildHalu (left) and Bios (right, \Cref{tab:main_results}) against the average over the four general capability benchmarks (\Cref{tab:results_olmes}). Up and to the right is better. All models lie within a 2.1-point band on the OLMES average, while FActScore varies by 10 (WildHalu) and 43 (Bios) points.}
\label{fig:tradeoff_general}
\end{figure*}

\Cref{fig:tradeoff_general} plots FActScore against the OLMES average for all OASST1 models, complementing the discussion in \Cref{subsec:general_tasks}.

\section{UnknownBench Precision--Recall Trade-off}
\label{sec:unknownbench_pr_appendix}

\begin{figure*}[t]
\centering
\iflatexml
\htmlfig{tradeoff_unknownbench}{Three scatter plots of refusal precision against recall on the FalseQA, NEC and RefuNQ subtasks, with iso-F1 contours. Recall Rewrite trades precision for substantially higher recall than the baselines.}%
\else
\begin{tikzpicture}
\begin{groupplot}[
  tradeoff, width=4.2cm, height=3.6cm,
  group style={group size=3 by 1, horizontal sep=0.9cm, ylabels at=edge left},
  ylabel={Precision}, xlabel={Recall},
  tolegend,
]
\nextgroupplot[legend to name=tradeoffUnknownLegend, title={FalseQA}, xmin=52, xmax=74, ymin=64, ymax=86]
\addplot[isof, domain=52:74, restrict y to domain=64:86] {60*x/(2*x-60)} node[isolabel, pos=1, anchor=south west] {F1=60};
\addplot[isof, domain=52:74, restrict y to domain=64:86] {65*x/(2*x-65)} node[isolabel, pos=1, anchor=south west] {65};
\addplot[isof, domain=52:74, restrict y to domain=64:86] {70*x/(2*x-70)} node[isolabel, pos=1, anchor=south east] {70};
\addplot[pt/base, tolabel] table[x=x, y=y, meta=label] {
x y label anchor
56.6 83.1 SFT south
55.5 82.4 FLAME north
59.1 77.8 \unitcut{} west
};
\addlegendentry{Baselines}
\addplot[pt/er, tolabel] table[x=x, y=y, meta=label] {
x y label anchor
54.8 83.5 ER east
58.1 82.8 {ER w/o brainst.} west
};
\addlegendentry{Evidence Rewrite (ours)}
\addplot[pt/rr, tolabel] table[x=x, y=y, meta=label] {
x y label anchor
64.1 74.0 RR west
};
\addlegendentry{Recall Rewrite (ours)}
\addplot[pt/rrabl, tolabel] table[x=x, y=y, meta=label] {
x y label anchor
71.2 67.3 0\% north
66.6 70.5 50\% north
69.2 71.7 100\% south
};
\addlegendentry{Recall Rewrite, \%Known ablation (\Cref{tab:cr_ablations})}
\nextgroupplot[title={NEC}, xmin=34, xmax=82, ymin=61, ymax=82]
\addplot[isof, domain=34:82, restrict y to domain=61:82] {55*x/(2*x-55)} node[isolabel, pos=1, anchor=south west] {F1=55};
\addplot[isof, domain=34:82, restrict y to domain=61:82] {65*x/(2*x-65)} node[isolabel, pos=1, anchor=south west] {65};
\addplot[isof, domain=34:82, restrict y to domain=61:82] {75*x/(2*x-75)} node[isolabel, pos=0.55, anchor=south west] {75};
\addplot[pt/base, tolabel] table[x=x, y=y, meta=label] {
x y label anchor
43.9 79.0 SFT south
43.9 77.7 FLAME west
37.7 72.5 \unitcut{} north
};
\addplot[pt/er, tolabel] table[x=x, y=y, meta=label] {
x y label anchor
39.1 76.4 ER north
50.9 75.0 {ER w/o brainst.} west
};
\addplot[pt/rr, tolabel] table[x=x, y=y, meta=label] {
x y label anchor
66.3 71.4 RR south
};
\addplot[pt/rrabl, tolabel] table[x=x, y=y, meta=label] {
x y label anchor
42.1 64.0 0\% north
42.1 67.4 50\% west
78.4 70.2 100\% north
};
\nextgroupplot[title={RefuNQ}, xmin=46, xmax=90, ymin=60, ymax=70]
\addplot[isof, domain=46:90, restrict y to domain=60:70] {60*x/(2*x-60)} node[isolabel, pos=1, anchor=south west] {F1=60};
\addplot[isof, domain=46:90, restrict y to domain=60:70] {65*x/(2*x-65)} node[isolabel, pos=1, anchor=south west] {65};
\addplot[isof, domain=46:90, restrict y to domain=60:70] {70*x/(2*x-70)} node[isolabel, pos=1, anchor=south west] {70};
\addplot[pt/base, tolabel] table[x=x, y=y, meta=label] {
x y label anchor
50.7 67.7 SFT south
53.1 67.8 FLAME {north west}
61.7 65.9 \unitcut{} north
};
\addplot[pt/er, tolabel] table[x=x, y=y, meta=label] {
x y label anchor
49.3 67.2 ER north
59.6 68.1 {ER w/o brainst.} south
};
\addplot[pt/rr, tolabel] table[x=x, y=y, meta=label] {
x y label anchor
79.1 62.6 RR north
};
\addplot[pt/rrabl, tolabel] table[x=x, y=y, meta=label] {
x y label anchor
51.0 65.5 0\% north
54.4 64.6 50\% north
85.8 61.9 100\% north
};
\end{groupplot}
\node[anchor=north] at ($(group c1r1.south west)!0.5!(group c3r1.south east) + (0,-0.8cm)$) {\pgfplotslegendfromname{tradeoffUnknownLegend}};
\end{tikzpicture}
\fi
\caption{Refusal precision vs.\ recall on the three UnknownBench subtasks (\Cref{tab:refusal_benchmark}). Each trained model yields a single operating point since refusal is a discrete generation decision; gray contours are iso-F1 lines. Hollow diamonds connect the Recall Rewrite \%Known ablation models (\Cref{tab:cr_ablations}). Recall Rewrite and the 100\%-known model trade precision for substantially higher recall than all baselines.}
\label{fig:tradeoff_unknownbench}
\end{figure*}
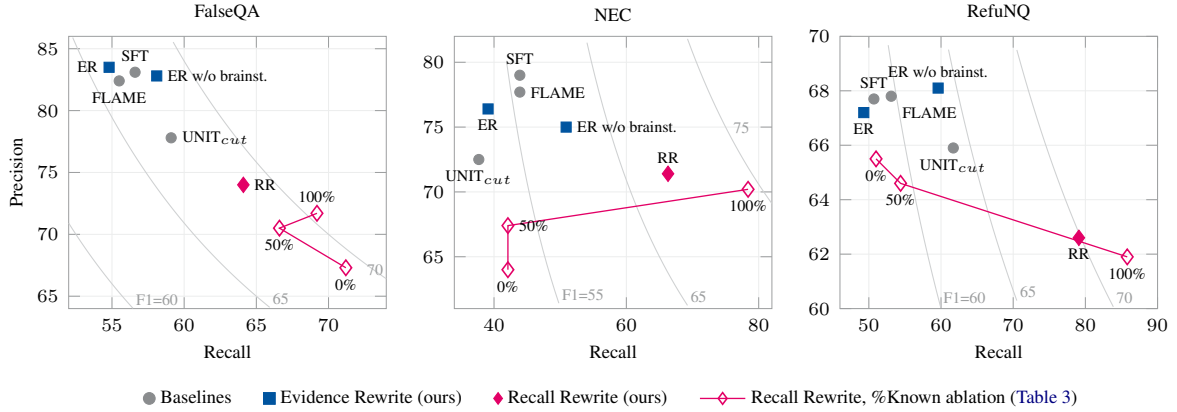

Since refusal is a discrete generation decision, our models do not expose an inference-time confidence threshold.
Each trained model corresponds to a single precision--recall operating point on UnknownBench, and its conservativeness is instead determined at training time by the strength of knowledge alignment.
\Cref{fig:tradeoff_unknownbench} therefore plots the results of \Cref{tab:refusal_benchmark} together with the \%Known ablation models of \Cref{tab:cr_ablations} against iso-F1 contours.

On all three subtasks, Recall Rewrite and its 100\%-known ablation trade precision for substantially higher recall than the baselines, and the 100\%-known model attains the highest F1 on every subtask (70.4 / 74.1 / 71.9).
The ablation models trained with a larger share of unknown claims (0\% and 50\%) fall back to baseline-level recall on NEC and RefuNQ at a lower precision, i.e., they refuse less reliably and less selectively.

The filter thresholds of \Cref{tab:cr_filters} provide a finer-grained control over this operating point.

\newcommand{\qexrule}{\noindent\rule{\columnwidth}{0.6pt}\par}
\newcommand{\qexhead}[1]{\addvspace{3pt}\noindent\textit{#1}\par\nobreak\noindent}
\newcommand{\qex}[9]{%
\par\addvspace{8pt plus 2pt minus 2pt}
\Needspace*{7\baselineskip}
\begingroup\footnotesize
\qexrule\nobreak\addvspace{2pt}
\noindent\refstepcounter{qexample}\label{#1}\textbf{Example \theqexample{} (#2): #3}\par\nobreak
\noindent\textit{Prompt:} #4\par
\qexhead{#5}#6\par
\qexhead{#7}#8\par
\addvspace{3pt}\noindent #9\par
\nobreak\addvspace{2pt}\qexrule
\endgroup
\addvspace{8pt plus 2pt minus 2pt}
}
\newcommand{\qref}{I cannot answer as I have insufficient information.}
\newcommand{\qecu}[3]{(E#1\,/\,C#2\,/\,U#3)}
\newcommand{\qprobing}{\par\addvspace{2pt}\noindent\textit{Probing:} }
\newcommand{\qq}[1]{Q: \enquote{#1} $\rightarrow$ }
\newcommand{\qind}{\hspace*{1.9em}}            %
\newcommand{\qfence}{\textasciigrave\textasciigrave\textasciigrave}

\section{Qualitative Analysis of Recall Rewrite}
\label{sec:qualitative_appendix}

This appendix complements the aggregate results with examples of where Recall Rewrite succeeds and where it fails.
We look at the three stages at which the method acts: the rewriting of the OASST1 training data (\Cref{subsec:qual_training}), the generations of the resulting model on WildHalu and Bios (\Cref{subsec:qual_generations}), and the regressions on IFEval and GSM8K (\Cref{subsec:qual_olmes}).
All examples come from the \qwenbaseid{} pipeline and from the Recall Rewrite and standard SFT models trained on OASST1 (\Cref{tab:main_results}).
Each example is shown side by side with the original and the rewritten response (\Cref{subsec:qual_training}) or with the responses of standard SFT and Recall Rewrite (\Cref{subsec:qual_generations,subsec:qual_olmes}), followed by the information that explains the outcome.

\subsection{Training Data Rewrites}
\label{subsec:qual_training}

\Cref{qex:darkmatter,qex:woodchuck,qex:prudence,qex:warsaw,qex:chatgpt,qex:haddaway,qex:rust,qex:pizza,qex:tribonacci} show original OASST1 responses next to their rewritten versions together with the claims that the pipeline classified as unknown.
For each removed claim we report how many of the ten probe answers ($J{=}5$ questions with $K{=}2$ answers each) entail (E), contradict (C), or are unrelated to (U) the claim.
Where it explains the outcome, a \textit{Probing} line additionally shows representative probing questions with the sampled base-model answers and their entailment label.

\paragraph{Intended behavior.}
In the intended case (\Cref{qex:darkmatter,qex:woodchuck}), the base model consistently contradicts one specific claim of the gold response, such as the share of dark matter in the universe or the daily food intake of a groundhog, and the rewriter removes exactly this claim with a local edit while leaving the rest of the response untouched.
Both removed claims are true.
Recall Rewrite removes them not because they are wrong, but because \qwenbaseid{} does not recall them: for dark matter, all ten probe answers give 27\%, which is the share of dark matter in the total mass--energy of the universe rather than in its matter.
\Cref{qex:prudence} shows the same mechanism leading to a refusal.
The gold response correctly attributes the lyrics to \enquote{Dear Prudence} by The Beatles, but in all ten probe answers the base model attributes them to \enquote{Beautiful Day} by U2.
Both content-bearing claims are therefore removed and the rewriter returns the refusal template.

\qex{qex:darkmatter}%
{success}%
{A single unknown quantitative claim is removed}%
{what is dark matter}%
{Gold response $R$}%
{Dark matter is an hypothetical form of matter, which does not interact with the electromagnetic field. It is thought to account for approximately 85\% of the matter in the known universe, if it does exist.}%
{Rewritten response $R^*$}%
{Dark matter is an hypothetical form of matter, which does not interact with the electromagnetic field.}%
{\textit{Removed claim:} \enquote{Dark matter is thought to account for approximately 85 percent of the matter in the known universe.} \qecu{0}{10}{0}. All seven retained knowledge-dependent claims are entailed by at least eight of the ten answers.
\qprobing \qq{Approximately what percentage of the matter in the known universe is believed to consist of dark matter?} \enquote{It is believed that about 27\% of the matter in the known universe consists of dark matter.} (C); \enquote{Dark matter is believed to make up around 27\% of the matter in the universe, while ordinary matter accounts for about 5\%.} (C). All ten answers to the five questions state 27\%.}

\qex{qex:woodchuck}%
{success}%
{A removed quantity is paraphrased consistently throughout the response}%
{How much wood would a woodchuck chuck if a woodchuck could chuck wood?}%
{Gold response $R$}%
{Since woodchucks (also known as Groundhogs) are pretty small animals, the amount of wood they could chuck would likely be part of their normal food consumption.\newline
Adult Groundhogs can eat up to one pound of vegetation per day. So, assuming their metabolism doesn't change:\newline
A woodchuck would chuck 1 pound of wood if a woodchuck could chuck wood.}%
{Rewritten response $R^*$}%
{Since woodchucks (also known as Groundhogs) are pretty small animals, the amount of wood they could chuck would likely be part of their normal food consumption.\newline
They consume a modest amount of vegetation each day. So, assuming their metabolism doesn't change:\newline
A woodchuck would chuck a modest amount of wood if a woodchuck could chuck wood.}%
{\textit{Removed claim:} \enquote{Adult groundhogs can eat up to one pound of vegetation per day.} \qecu{0}{10}{0}.}

\qex{qex:prudence}%
{success}%
{The base model consistently recalls a different answer, so the response becomes a refusal}%
{[lyrics] of the verse go like this: \enquote{The sun is up, the sky is blue, it's beautiful and so are you.} Which song is that and who is the author of the [lyrics]?}%
{Gold response $R$}%
{Those are part of the lyrics of the song \enquote{Dear Prudence} by The Beatles, written by John Lennon.}%
{Rewritten response $R^*$}%
{\qref}%
{\textit{Removed claims:} \enquote{The lyrics \textelp{} appear in the song `Dear Prudence'.} \qecu{0}{10}{0}; \enquote{John Lennon wrote the lyrics of the song `Dear Prudence'.} \qecu{0}{7}{3}. \textit{Retained:} \enquote{`Dear Prudence' is a song performed by the Beatles.} \qecu{10}{0}{0}.
\qprobing \qq{Which song includes the lyric `The sun is up, the sky is blue, it's beautiful and so are you.'?} \enquote{The song `Beautiful Day' by U2 contains the lyric \textelp{}} (C); \enquote{The song is `Beautiful Day' by U2.} (C). All ten answers to the five questions name \enquote{Beautiful Day} by U2.}

\paragraph{Failure modes.}
\Cref{qex:warsaw,qex:chatgpt,qex:haddaway,qex:rust} show the four failure modes we observed most frequently.
First, \emph{the probing questions do not target the information that the claim adds}.
In \Cref{qex:warsaw}, the claim to be tested contributes one piece of new information to the response, the year 1596, since Warsaw as the capital is asserted (and classified as known) by a separate claim.
All five generated questions nevertheless ask for the city \emph{and} the year.
The base model recalls the year in nine of the ten answers, but three questions receive a wrong city (Lwów, Kraków, Lublin), which counts as three contradicting questions and removes the claim, whereas the near-identical retained claim \enquote{The capital \textelp{} moved from Kraków to Warsaw in 1596} passes because its questions happen to name the city.
Untargeted questions thus make the classification depend on which part of a compound question the base model gets wrong, and can classify claims carrying the same information differently.
Overall, 392 of the 8,199 removed claims (4.8\%) are entailed by at least six of the ten answers.
Second, \emph{the probes fail to elicit related answers}.
1,575 of the 8,199 removed claims (19.2\%) are removed without a single contradicting answer, i.e., only because too few of the sampled answers were judged to entail the claim.
\Cref{qex:chatgpt} is such a case: the answers describe \enquote{language models} or \enquote{NLP models}, which the entailment check does not accept as evidence for a claim about \emph{large} language models, so the classification reflects the strictness of the probe and entailment step rather than the knowledge of the base model.
Third, \emph{the rewriter removes more than the unknown claim}.
In the same example, only this generic definition is classified as unknown, yet the rewriter also drops the sentence on ChatGPT's capabilities although all of its claims were retained.
Fourth, \emph{the rewritten response loses coherence or changes the content}.
After removing all references to the song in \Cref{qex:haddaway}, the remaining clarification question still asks about \enquote{the song}, although the response no longer says which song is meant.
In \Cref{qex:rust}, the rewriter reacts to the removal of a claim about \texttt{match} expressions by rewriting the code itself into an \texttt{if}/\texttt{else} chain and dropping the code fence.
This last case is specific to code and other structured responses, where many claims describe how the response is constructed rather than what it states.
Over-truncation is nonetheless rare in the non-refusal set: the median rewritten response retains 97\% of the characters of the original, and 1,127 of the 3,165 responses have no claim removed at all.

\qex{qex:warsaw}%
{failure}%
{The probing questions test the city rather than the year that the claim adds}%
{What was the capital of Poland in 1640?}%
{Gold response $R$}%
{Warsaw.\newline
It became the capital of Poland-Lithuania Commonwealth in 1596.\newline
It was chosen as the capital because it was the mid-point between the then Polish capital Cracow, and Lithuanian capital Vilnius.}%
{Rewritten response $R^*$}%
{Warsaw.\newline
It was chosen as the capital because it was the mid-point between the then Polish capital Cracow, and Lithuanian capital Vilnius.}%
{\textit{Removed claim:} \enquote{Warsaw became the capital of the Polish--Lithuanian Commonwealth in 1596.} \qecu{7}{3}{0}. \textit{Retained, among others:} \enquote{Warsaw was the capital of Poland in 1640.} \qecu{8}{2}{0}; \enquote{The capital of the Polish--Lithuanian Commonwealth moved from Kraków to Warsaw in 1596.} \qecu{4}{0}{6}.
\qprobing All five questions ask for city and year, e.g., \qq{Name the city that assumed the role of capital for the Polish--Lithuanian Commonwealth and state the year when this happened.} \enquote{Lwów (now Lviv, Ukraine) became the capital of the Polish--Lithuanian Commonwealth in 1596.} (C); \enquote{The city of Warsaw became the capital of the Polish--Lithuanian Commonwealth in 1596.} (E). Nine of the ten answers give the year 1596; the three contradictions (Lwów, Kraków, Lublin) all concern the city. The retained near-duplicate claim was probed with questions that name Kraków as the previous capital, e.g., \qq{Identify the city that replaced Kraków as the seat of the Polish--Lithuanian Commonwealth and state the year the replacement happened.} \enquote{The city that replaced Kraków \textelp{} was Warsaw. The replacement happened in 1596.} (E).}

\qex{qex:chatgpt}%
{failure}%
{No question entails a generic claim, and the rewriter removes more than the unknown claim}%
{Do you know of Chat GPT?}%
{Gold response $R$}%
{ChatGPT is a large language model (LLM) developed by OpenAI, based on the GPT-3 architecture. It is designed to understand and respond to natural language input from users, and its capabilities include tasks such as language translation, answering questions, and generating human-like responses to prompts.}%
{Rewritten response $R^*$}%
{Yes. ChatGPT is developed by OpenAI and is based on the GPT-3 architecture.}%
{\textit{Removed claim:} \enquote{A large language model is trained to process and generate human language.} \qecu{2}{0}{8}. \textit{Retained, among others:} \enquote{ChatGPT is a large language model.} \qecu{6}{0}{4}; \enquote{ChatGPT can perform language translation.} \qecu{10}{0}{0}; \enquote{ChatGPT is designed to respond to natural language input from users.} \qecu{9}{0}{1}.
\qprobing \qq{Which type of AI model is trained to analyze and produce natural human language?} \enquote{A language model is an AI model specifically designed to understand and generate natural human language.} (U); \enquote{A natural language processing (NLP) model is trained to understand, interpret, and generate human language.} (U). Eight of the ten answers name language or NLP models without the qualifier \emph{large} and are labeled unrelated.}

\qex{qex:haddaway}%
{failure}%
{The rewritten response is no longer coherent}%
{What is love? Baby don't hurt me}%
{Gold response $R$}%
{Sorry, could you please clarify your task? Would you like to know more about the song 1993 song \enquote{What is Love} from Haddaway?}%
{Rewritten response $R^*$}%
{Sorry, could you please clarify your task? Would you like to know more about the song?}%
{\textit{Removed claims:} \enquote{The phrase `What is love? Baby don't hurt me' is a lyric from the song `What Is Love'.} \qecu{3}{4}{3}; \enquote{The song `What Is Love' was released in 1993.} \qecu{1}{9}{0}; \enquote{The song `What Is Love' was performed by the artist Haddaway.} \qecu{0}{10}{0}.}

\qex{qex:rust}%
{failure}%
{The rewriter changes the code instead of removing a claim}%
{I've been learning Rust for a few weeks now, and have been trying my hand at the fizzbuzz problem. [\emph{code of} \texttt{fn fizzbuzz<'a>(n: usize) -> \&'a str} \emph{using a} \texttt{match} \emph{expression}] For some reason, I get a compiler error that I'm struggling to understand. [\texttt{error[E0515]: cannot return value referencing temporary value} \textelp{}] Could you please explain to my why this happens and how I can fix it?}%
{Gold response $R$}%
{The simplest way to solve this is to just return a \texttt{String} rather than a \texttt{\& str}:\newline
\texttt{\qfence rust}\newline
\texttt{fn fizzbuzz(n: usize) -> String \{}\newline
\qind\texttt{match (n \% 3, n \% 5) \{}\newline
\qind\qind\texttt{(0, 0) => "fizzbuzz".to\_string(),}\newline
\qind\qind\texttt{(0, \_) => "fizz".to\_string(),}\newline
\qind\qind\texttt{(\_, 0) => "buzz".to\_string(),}\newline
\qind\qind\texttt{\_ => n.to\_string(),}\newline
\qind\texttt{\}}\newline
\texttt{\}}\newline
\texttt{\qfence}}%
{Rewritten response $R^*$}%
{The simplest way to solve this is to just return a \texttt{String} rather than a \texttt{\& str}:\newline
\texttt{fn fizzbuzz(n: usize) -> String \{}\newline
\qind\texttt{if n \% 3 == 0 \&\& n \% 5 == 0 \{}\newline
\qind\qind\texttt{"fizzbuzz".to\_string()}\newline
\qind\texttt{\} else if n \% 3 == 0 \{}\newline
\qind\qind\texttt{"fizz".to\_string()}\newline
\qind\texttt{\} else if n \% 5 == 0 \{}\newline
\qind\qind\texttt{"buzz".to\_string()}\newline
\qind\texttt{\} else \{}\newline
\qind\qind\texttt{n.to\_string()}\newline
\qind\texttt{\}}\newline
\texttt{\}}}%
{\textit{Removed claim:} \enquote{All arms of a match expression must produce values of the same type.} \qecu{7}{3}{0}. All nine other knowledge-dependent claims (on \texttt{\&str}, \texttt{String}, lifetimes, and temporaries) were retained. The prompt is abbreviated.}

\paragraph{Over-refusals.}
The rewriter returns the refusal template for 302 of the 3,467 OASST1 examples.
In 7 of these no claim was removed at all (\Cref{qex:pizza}), and in 114 cases at most a quarter of the knowledge-dependent claims were removed.
Such refusals are over-represented for code: 95 of the 302 refusals (31\%) but only 434 of the 3,165 non-refusals (14\%) contain code, and 64 of the 114 refusals with few removed claims are code answers.
\Cref{qex:tribonacci} is typical.
A single claim about the index range is classified as unknown, and its contradicting answers reflect the ambiguity of the claim (indexing from 0 or 1, negative extensions of the sequence) rather than missing knowledge.
The code itself is unaffected, but the rewriter judges the remaining information insufficient rather than returning the unchanged code.
These refusals originate from the insufficient-information decision of the rewriter (\Cref{fig:template_redaction}) rather than from the knowledge classification.
Constraining this decision, for instance by never refusing when no claim was removed, is a straightforward improvement for future work.

\qex{qex:pizza}%
{over-refusal}%
{No claim was removed}%
{All pizzas are food. I have not eaten any food today. Have I eaten any pizzas today?}%
{Gold response $R$}%
{No. You stated you have not eaten any food today, and since all pizzas are food, it can be deduced that you have not eaten any pizzas today.}%
{Rewritten response $R^*$}%
{\qref}%
{\textit{Removed claims:} none. Four of the five claims are non-knowledge-dependent (logical inferences over the prompt) and the fifth, \enquote{If a conditional statement `if P then Q' is true and Q is false, then P is false.} \qecu{7}{0}{3}, is known.}

\qex{qex:tribonacci}%
{over-refusal}%
{One of 16 knowledge-dependent claims is removed, the code is unaffected}%
{Write Python code to generate N Tribonacci numbers}%
{Gold response $R$}%
{\texttt{\qfence python}\newline
\texttt{def tribonacci(n):}\newline
\qind\texttt{if n == 0 or n == 1:}\newline
\qind\qind\texttt{return 0}\newline
\qind\texttt{if n == 2:}\newline
\qind\qind\texttt{return 1}\newline
\qind\texttt{return tribonacci(n-1) + tribonacci(n-2)}\newline
\qind\qind\qind\texttt{+ tribonacci(n-3)}\newline
\texttt{\qfence}}%
{Rewritten response $R^*$}%
{\qref}%
{\textit{Removed claim:} \enquote{A Tribonacci index is a non-negative integer.} \qecu{6}{4}{0}. \textit{Retained, among others:} \enquote{For every integer $n \geq 3$, the Tribonacci number at index $n$ equals the sum of the Tribonacci numbers at indices $n{-}1$, $n{-}2$, and $n{-}3$.} \qecu{7}{0}{3}; \enquote{In Python, the keyword def begins a function definition.} \qecu{10}{0}{0}; \enquote{In Python, indentation determines which statements belong to a block.} \qecu{10}{0}{0}.
\qprobing \qq{What type of number is used to label positions in the Tribonacci sequence (i.e., what kind of number can a Tribonacci index be)?} \enquote{A Tribonacci index can be any positive integer, as it represents the position in the sequence starting from 1.} (C); \enquote{A Tribonacci index can be any non-negative integer \textelp{}} (E). \qq{Which category of numbers describes valid indices for the Tribonacci sequence (for example: negative numbers, positive only, or including zero)?} \enquote{Valid indices for the Tribonacci sequence include both positive and negative numbers, as well as zero, since the sequence can be extended infinitely in both directions.} (C). The last code line is wrapped for layout.}

\subsection{Model Generations}
\label{subsec:qual_generations}

\begin{table}[H]
\centering
\footnotesize
\setlength{\tabcolsep}{3pt}
\begin{tabular}{lrr}
\toprule
 & WildHalu & Bios \\
\midrule
Recall Rewrite (RR) refuses & 55 & 252 \\
\quad std.\ SFT $<40\%$ true claims & 31 & 224 \\
\quad std.\ SFT 40--70\% true claims & 15 & 23 \\
\quad std.\ SFT $\geq 70\%$ true claims & 8 & 1 \\
\quad std.\ SFT also refuses & 1 & 4 \\
Std.\ SFT refuses, RR answers & 1 & 0 \\
\addlinespace
Both models answer & 444 & 248 \\
\quad RR has fewer false claims & 265 & 215 \\
\quad same number of false claims & 101 & 11 \\
\quad RR has more false claims & 78 & 22 \\
\addlinespace
\multicolumn{3}{@{}l}{Avg.\ true\,/\,false claims when both answer} \\
\quad std.\ SFT & 17.1\,/\,4.5 & 14.7\,/\,15.6 \\
\quad RR & 13.8\,/\,2.6 & 10.3\,/\,8.0 \\
\bottomrule
\end{tabular}
\caption{Per-prompt comparison of Recall Rewrite and standard SFT on the 500 WildHalu and 500 Bios prompts, based on the fact-check verdicts (true / false / not known) of \Cref{subsec:fact-checking}. The share of true claims is computed over all claims of the standard SFT response. Refusals are identified with the evaluation pipeline's LLM refusal detector, so the refusal counts match the \#Ref.\ column of \Cref{tab:main_results}; the true-claim buckets cover the prompts on which standard SFT answers.}
\label{tab:qual_gen_stats}
\end{table}

\Cref{tab:qual_gen_stats} categorizes the 500 WildHalu and 500 Bios prompts by the behavior of the two models.
\Cref{qex:jager,qex:lovenox,qex:brandenburg,qex:monocle,qex:glaceon} show representative generations with the number of true\,/\,false\,/\,not known claims according to the fact-checking pipeline of \Cref{subsec:fact-checking}.

\paragraph{Refusals.}
Recall Rewrite refuses on 252 Bios prompts.
Standard SFT answers on 248 of these prompts; for 224 (90\%), fewer than 40\% of its claims for the same entity are true, and only for one entity (Luiz Inácio Lula da Silva, 39 true and 13 false claims) is the standard SFT response predominantly correct.
Bios refusals thus almost exclusively concern entities on which standard SFT hallucinates, such as Maja Jager in \Cref{qex:jager}, where standard SFT invents a Swedish singer while the entity is a Danish archer.
The 40\% threshold, if anything, overstates the quality of standard SFT, since the fact-checker marks generic claims such as \enquote{she has released several albums} as true even when the response describes a fictitious person.
On WildHalu, whose entities are less long-tail (cf.\ the \%Supp.\ of standard SFT in \Cref{tab:main_results}), the picture is more mixed.
Of the 55 refusals, 31 concern entities on which standard SFT is mostly wrong, but 8 concern entities on which it is mostly right.
For Lovenox (\Cref{qex:lovenox}), for example, all 21 claims of the standard SFT response are supported.
These over-refusals are the price of the higher \%Supp.\ (\Cref{tab:main_results}) and the higher refusal recall (\Cref{tab:refusal_benchmark}).

\qex{qex:jager}%
{correct refusal}%
{Standard SFT describes a fictitious person}%
{Tell me a bio of Maja Jager.}%
{Standard SFT}%
{Maja Jager is a Swedish singer and songwriter who was born on March 17, 1989 in Stockholm, Sweden. She began her music career at a young age, singing in church choirs and participating in various music competitions.\newline
Jager gained widespread recognition in 2013 after winning the Swedish version of the reality TV show \enquote{The Voice}. She went on to represent Sweden in the Eurovision Song Contest that same year, where she finished in 10th place.\newline
Since then, Jager has released several albums and singles, including her debut album \enquote{Maja Jager} in 2014 and her second album \enquote{The Voice} in 2016. She has also collaborated with various artists, including fellow \enquote{The Voice} contestant Marcus \& Martinus.\newline
In addition to her music career, Jager is also known for her humanitarian work, including her involvement with the Swedish Red Cross and her efforts to raise awareness for mental health issues.}%
{Recall Rewrite}%
{\qref}%
{\textit{Fact-check (true\,/\,false\,/\,not known):} standard SFT 1\,/\,21\,/\,0. Wikipedia: \enquote{Maja Buskbjerg Jager (born 22 December 1991) is a Danish recurve archer.}}

\qex{qex:lovenox}%
{over-refusal}%
{Standard SFT is fully correct}%
{In a paragraph, could you tell me what you know about Lovenox?}%
{Standard SFT}%
{Lovenox is the brand name for enoxaparin, a low molecular weight heparin (LMWH) used to prevent and treat blood clots. It is commonly used to prevent deep vein thrombosis (DVT) and pulmonary embolism (PE) in patients who are at risk, such as those who have undergone surgery or have been bedridden for an extended period of time. It is also used to treat existing blood clots in the legs or lungs, and to prevent recurrent clots. Lovenox works by inhibiting the formation of blood clots, thereby reducing the risk of clot-related complications. It is administered via subcutaneous injection and is generally well-tolerated, although like all medications, it can cause side effects such as bleeding, bruising, and injection site reactions. As with any medication, Lovenox should only be used under the guidance of a healthcare provider.}%
{Recall Rewrite}%
{\qref}%
{\textit{Fact-check:} standard SFT 21\,/\,0\,/\,0.}

\paragraph{Both models answer.}
When both models answer, Recall Rewrite produces shorter responses with fewer claims (WildHalu: 13.8 true and 2.6 false claims per response vs.\ 17.1 and 4.5 for standard SFT; Bios: 10.3 and 8.0 vs.\ 14.7 and 15.6) and has fewer false claims than standard SFT for 265 of the 444 WildHalu and 215 of the 248 Bios prompts.
In the best cases it drops exactly the wrong details (\Cref{qex:brandenburg}) or replaces a fabricated description with a correct one (\Cref{qex:monocle}).
Recall Rewrite is not immune to hallucination, however: it produces more false claims than standard SFT on 78 WildHalu and 22 Bios prompts.
\Cref{qex:glaceon} shows such a case.
The model answers a question about an entity it evidently does not know instead of refusing, and fabricates a detailed description whose errors merely differ from those of standard SFT.
The training thus reduces how often the model guesses, but when it does guess, its output is no more reliable.

\qex{qex:brandenburg}%
{correction}%
{Recall Rewrite drops exactly the wrong details}%
{In a paragraph, could you tell me what you know about Brandenburg Gate?}%
{Standard SFT}%
{The Brandenburg Gate is a neoclassical monument located in Berlin, Germany. It was built between 1788 and 1793 to commemorate the Peace of Westphalia, which ended the War of the First Coalition against France. The gate is made of white marble and features a large arch with four fluted columns supporting a triangular pediment. It has been a symbol of German unity and freedom for centuries, and has been the site of many important historical events, such as the proclamation of German reunification in 1990. Today, it is one of the most iconic landmarks in Berlin, and a popular tourist attraction.}%
{Recall Rewrite}%
{The Brandenburg Gate is a neoclassical monument in Berlin, Germany. It was built between 1788 and 1791 and is one of the most iconic landmarks in the city. The gate is a symbol of the unity of Germany and has been the site of many important historical events. It has been the setting for numerous political rallies and demonstrations, including the German Unification Day celebrations in 1990.}%
{\textit{Fact-check:} standard SFT 7\,/\,6\,/\,2 (false: construction years, Peace of Westphalia, white marble, four fluted columns); Recall Rewrite 11\,/\,0\,/\,0.}

\qex{qex:monocle}%
{correction}%
{A fabricated description is replaced by a correct one}%
{In a paragraph, could you tell me what you know about Monocle3?}%
{Standard SFT}%
{Monocle3 is a deep brain stimulation device that is used to treat movement disorders such as Parkinson's disease. It is a small, implantable device that is placed in the chest and connected to electrodes that are implanted in specific areas of the brain. The device delivers electrical impulses to the brain to help regulate abnormal brain activity that can cause symptoms such as tremors, rigidity, and bradykinesia. Monocle3 is designed to be programmable, allowing doctors to adjust the settings based on the patient's individual needs. It is a wireless device that is powered by a battery, which can be replaced during a simple surgical procedure. Monocle3 is intended to be a long-term solution for patients with movement disorders, and it has been shown to be effective in reducing symptoms and improving quality of life for many patients.}%
{Recall Rewrite}%
{Monocle3 is a computational tool designed for the analysis of single-cell RNA sequencing (scRNA-seq) data. It provides a suite of algorithms for processing and analyzing scRNA-seq data, including quality control, normalization, dimensionality reduction, and clustering. Monocle3 also includes tools for visualizing the results of scRNA-seq analysis, such as trajectory inference and pseudotime analysis. The tool is implemented in R and provides a user-friendly interface for performing these analyses. Overall, Monocle3 is a powerful and versatile tool for analyzing scRNA-seq data and gaining insights into the complex biology of single cells.}%
{\textit{Fact-check:} standard SFT 5\,/\,18\,/\,0 (the true claims are generic statements about Parkinson's disease); Recall Rewrite 15\,/\,0\,/\,3.}

\qex{qex:glaceon}%
{regression}%
{The model answers instead of refusing and fabricates the details}%
{In a paragraph, could you tell me what you know about Glaceon?}%
{Standard SFT}%
{Glaceon is a fictional Pokémon introduced in the anime series Pokémon the Series: XY. It is a Water/Ice-type Pokémon that evolves from Wailord at level 40. In the anime, Glaceon is known for its ability to control ice and its association with the winter season. It is also depicted as a gentle and peaceful Pokémon that is often used by trainers for its defensive capabilities. In the Pokémon games, Glaceon is known for its unique design and its ability to learn a variety of powerful moves, making it a formidable opponent in battles. Overall, Glaceon is a popular and beloved Pokémon that is cherished by fans for its unique characteristics and abilities.}%
{Recall Rewrite}%
{Glaceon is a Pokémon introduced in Generation VII. It is a dual-type Ice/Flying Pokémon. In its base form, Glaceon has white and blue coloration with icy blue streaks running down its sides. It has small, round eyes with a white ring around them, and a small, pointed snout. Its wings are a light blue color and it has a small, triangular tail. In its evolved forms, its coloration changes and it gains additional features. As a Pokémon, Glaceon is known for its speed and agility in the air.}%
{\textit{Fact-check:} standard SFT 15\,/\,4\,/\,4; Recall Rewrite 6\,/\,15\,/\,1. Both descriptions are largely fabricated (Glaceon is an Ice-type evolution of Eevee introduced in Generation IV, without wings); Recall Rewrite invents a physical appearance and happens to state more checkable false details.}

\subsection{IFEval and GSM8K Regressions}
\label{subsec:qual_olmes}

\begin{table}[H]
\centering
\footnotesize
\setlength{\tabcolsep}{2pt}
\begin{tabular}{lrr}
\toprule
 & GSM8K & IFEval \\
\midrule
\# prompts & 1,319 & 541 \\
Std.\ SFT correct, RR wrong & 102 & 71 \\
Std.\ SFT wrong, RR correct & 71 & 54 \\
\addlinespace
Starts with refusal template & 0\,/\,0 & 0\,/\,30 \\
Contains a refusal phrase & 0\,/\,0 & 2\,/\,37 \\
Acc.\ on the 30 refused prompts & -- & 66.7\,/\,13.3 \\
Acc.\ on prompts w/o refusal phrase & 83.6\,/\,81.3 & 56.5\,/\,56.3 \\
Avg.\ response length (tokens) & 117\,/\,113 & 273\,/\,209 \\
\bottomrule
\end{tabular}
\caption{Per-example comparison of standard SFT and Recall Rewrite on GSM8K (exact match) and IF\discretionary{-}{}{}Eval (prompt-level loose accuracy). Pairs are standard SFT\,/\,Recall Rewrite (RR); the first three rows count prompts, the remaining rows are counts of responses, accuracies in \%, and tokens. \enquote{Refused prompts} are those where the Recall Rewrite response starts with the refusal template, \enquote{prompts without refusal phrase} the 504 IFEval prompts (all 1,319 GSM8K prompts) whose Recall Rewrite response contains no refusal phrase.}
\label{tab:qual_olmes_stats}
\end{table}

\Cref{tab:qual_olmes_stats} contrasts the per-example predictions of the two models on the two OLMES tasks with the largest differences in \Cref{tab:results_olmes}.

\paragraph{GSM8K.}
The 2.3-point drop results from 102 problems that standard SFT solves and Recall Rewrite does not, against 71 problems in the opposite direction.
None of the 1,319 responses of either model contains a refusal, every response ends in a numeric answer, and the average response length is unchanged.
The errors are ordinary reasoning slips such as omitting a step (\Cref{qex:gsm8k}), so we see no evidence that the refusal behavior affects GSM8K and attribute the difference to training variance.

\qex{qex:gsm8k}%
{GSM8K}%
{An omitted step, no refusal}%
{Ashley's pizza delivery costs \$15. What is the total amount that Ashley should give the delivery man if she wants to give a tip that is equal to 1/5 of the amount she ordered?}%
{Standard SFT (correct)}%
{The tip should be 15 / 5 = 3 dollars. So the total amount should be 15 + 3 = 18 dollars. So the answer is 18.}%
{Recall Rewrite (wrong)}%
{The tip should be 15 / 5 = 3. So the answer is 3.}%
{Gold answer: 18 (exact match). Typical of the 102 GSM8K problems that only standard SFT solves: both models produce a chain of arithmetic steps and a numeric answer.}

\paragraph{IFEval.}
On IFEval, the picture is different.
Recall Rewrite begins its response with the refusal template for 30 of the 541 prompts and contains a refusal phrase in 37 responses (standard SFT: 0 and 2).
Its accuracy on these 30 prompts is 13.3\% compared to 66.7\% for standard SFT, whereas on the 504 prompts without any refusal phrase the two models are on par (56.3\% vs.\ 56.5\%).
Refusals therefore account for the entire IFEval difference.
The refused prompts are predominantly creative-writing tasks (jokes, poems, riddles, raps) and opinion questions rather than knowledge-seeking questions (\Cref{qex:ifeval_joke,qex:ifeval_essay}).
The 302 refusals in the training data include refusals to such prompts (e.g., \enquote{Tell me a funny joke!} or \enquote{Please create an ASCII drawing of a cat wearing a hat.}, cf.\ \Cref{subsec:qual_training}), which the model appears to generalize into a tendency to refuse creative tasks even though they require no factual knowledge.
Reducing over-refusals of this kind in the training data is therefore the most direct lever for closing the IFEval gap.

\qex{qex:ifeval_joke}%
{IFEval}%
{A creative-writing prompt is refused}%
{Write a short and funny joke about a guy who works at the IRS. Include at least one placeholder represented by square brackets.}%
{Standard SFT (correct)}%
{Sure, here is an example:\newline
Why did the guy who worked at the IRS make a good impression on his first day?\newline
Because he was [tax] ready!}%
{Recall Rewrite (wrong)}%
{\qref}%
{Instruction checked: \texttt{detectable\_content:\allowbreak number\_placeholders}. Recall Rewrite responds with the refusal template only.}

\qex{qex:ifeval_essay}%
{IFEval}%
{A knowledge-heavy writing prompt is refused}%
{Write a 200 word essay on the 2000 presidential election. The title should be wrapped in double angular brackets, i.e. <\kern0pt<title>\kern0pt>.}%
{Standard SFT (correct)}%
{<\kern0pt<The 2000 Presidential Election>\kern0pt>\newline
The 2000 presidential election was one of the most controversial and closely watched elections in US history. The election was between the incumbent president, George W. Bush, and Vice President Al Gore. The election was ultimately decided by the Supreme Court in Bush v. Gore, which ruled that the Florida vote recount was unconstitutional. \textelp{}}%
{Recall Rewrite (wrong)}%
{\qref}%
{Instruction checked: \texttt{detectable\_format:\allowbreak title}. IFEval only verifies the format, so the factual errors of the standard SFT response (Bush was not the incumbent) do not count against it. The standard SFT response is truncated at \textelp{}.}

\section{Prompts}
\label{sec:prompts_appendix}
The rest of the appendix lists the different prompts used for the different components in this work.

\begin{figure*}[!t]
    \centering
    \begin{minipage}{0.9\textwidth}

        \begin{tabular}{|l|p{10cm}|}
            \hline
            \textbf{Example prompt 1}
                & \begin{lstlisting}[basicstyle=\scriptsize\ttfamily,breaklines=true]
Here is a question from a user: "What are the principles at play in UHPLC-MS analysis?". To answer the above question, do you need the factual knowledge from Wikipedia? Give an answer using the format: "Answer: Yes or No".
                    \end{lstlisting}
                \\
            \hline

            \textbf{Example response 1}
                & \begin{lstlisting}[basicstyle=\scriptsize\ttfamily,breaklines=true]
Answer: Yes
                    \end{lstlisting}
                \\
            \hline

            \multicolumn{2}{|c|}{
                ...
            } \\
            \hline

            \textbf{Example prompt 10}
                & \begin{lstlisting}[basicstyle=\scriptsize\ttfamily,breaklines=true]
Here is a question from a user: "Is the internet's focus on engagement the root of most of its problems and shortcomings?". To answer the above question, do you need the factual knowledge from Wikipedia? Give an answer using the format: "Answer: Yes or No".
                    \end{lstlisting}
                \\
            \hline

            \textbf{Example response 10}
                & \begin{lstlisting}[basicstyle=\scriptsize\ttfamily,breaklines=true]
Answer: No
                    \end{lstlisting}
                \\
            \hline

            \textbf{Instruction}
                & \begin{lstlisting}[basicstyle=\scriptsize\ttfamily,breaklines=true]
Here is a question from a user: "{prompt}". To answer the above question, do you need the factual knowledge from Wikipedia? Give an answer using the format: "Answer: Yes or No".
                    \end{lstlisting}
                \\
            \hline
            \end{tabular}

\end{minipage}

\caption{Few-shot prompt for classifying instructions as knowledge-seeking
or non-knowledge-seeking (non-factual), used in the FLAME pipeline to determine whether to replace a gold response
with a base model generation. Examples are adapted from \citet{flame}.}
\label{fig:factual_prompt_classifier}

\end{figure*}

\begin{figure*}[!t]
    \centering
    \begin{minipage}{0.9\textwidth}

        \begin{lstlisting}[
            basicstyle=\scriptsize\ttfamily,
            breaklines=true,
            frame=single,
            breakindent=0pt,
            framesep=3mm,
        ]


You are trying to verify how factual a response to a question or request is. To do so, you need to break down a sentence and extract as many fine-grained facts mentioned in the response. Each of these fine-grained facts should be verifiable against reliable external world knowledge (e.g., via Wikipedia). Any story, personal experiences, hypotheticals (e.g., "would be" or subjunctive), subjective statements (e.g., opinions), suggestions, advice, instructions, and other such content should not be included in the list. Biographical, historical, scientific, and other such texts are not personal experiences or  stories. You should extract verifiable facts from them. Each fact should also be describing either one single event (e.g., "Nvidia is founded in 1993 in Sunnyvale, California, U.S.") or single state (e.g., "UMass Amherst has existed for 161 years.") with necessary time and location information. Quotations should be extracted verbatim with the source when available. Listed references should be ignored.

Extract fine-grained facts from the sentence between <SOS> and <EOS>. You should focus on the named entities and numbers in the sentence and extract relevant information from the sentence. Do not extract claims from the question. The question and other sentences are only context for you to recover pronouns, definite phrases (e.g., "the victims" or "the pope"), and so on. Each fact should be understandable on its own and require no additional context. This means that you need to always related the extracted claims to the question. This also means that all entities must be referred to by name but not pronoun. Use the name of entities rather than definite noun phrases (e.g., 'the teacher') whenever possible. If a definite noun phrase is used, be sure to add modifiers (e.g., a embedded clause, a prepositional phrase, etc.). Each fact must be situated within relevant temporal and location whenever needed. Keep each fact to one sentence with zero or at most one embedded clause. You do not need to justify what you extract.

If there is no verifiable fact in the sentence, please write "No verifiable claim."

Here are some examples:

Question: What NASA programs would support our college in starting a robotics program?
Response: NASA has several programs that can support colleges in starting a robotics program. Here are a few:
<SOS>1. NASA Robotics Alliance Project (RAP): This program provides educational resources and support for robotics teams, including college-level teams, that are participating in NASA robotics competitions.<EOS>
2. NASA Minority University Research and Education Project (MUREP): This program provides funding and resources for colleges and universities with a significant minority student population to develop research and education programs in STEM fields, including robotics.
3. NASA's Robotics Education Project: This project provides robotics education materials and resources for educators, including college-level educators, to use in their classrooms.
4. NASA's Space Technology Mission Directorate (STMD): This directorate funds research and development in advanced technologies, including robotics, that can support NASA's mission to explore space.
Sentence to be focused on: 1. NASA Robotics Alliance Project (RAP): This program provides educational resources and support for robotics teams, including college-level teams, that are participating in NASA robotics competitions.
Facts:
- NASA has a program called NASA Robotics Alliance Project (RAP).
- NASA Robotics Alliance Project provides educational resources for robotics teams.
- NASA Robotics Alliance Project provides supports for robotics teams.
- NASA Robotics Alliance Project provides supports for college-level teams that are participating in NASA robotics competitions.

[...]

Question: How do trees know when to stop growing?
Thanks everyone i learned a lot more about trees.(:
Response: <SOS>Ah yes, tomatoes, this is a big problem with tomato plants.<EOS>
Sentence to be focused on: Ah yes, tomatoes, this is a big problem with tomato plants.
Facts:
No verifiable claim.

[...]

Extract *verifiable atomic* fact.

{snippet}
Sentence to be focused on: {sentence}
Facts:


        \end{lstlisting}

\end{minipage}

\caption{Claim decomposition prompt from \citet{song-etal-2024-veriscore}, used to extract
verifiable atomic facts from model responses. Some few-shot examples omitted for brevity.}
\label{fig:claim_decomposition_prompt}

\end{figure*}

\begin{figure}[!t]

    \begin{lstlisting}[
        basicstyle=\scriptsize\ttfamily,
        linewidth=.95\linewidth,
        frame=single,
        breaklines=true,
        breakindent=0pt,
        framesep=3mm,
        numbers=none,
    ]
Answer the question about {topic} based on the given context.

Title: {retrieved_passages[0].title}
Text: {retrieved_passages[0].text}

Title: {retrieved_passages[1].title}
Text: {retrieved_passages[1].text}

Title: {retrieved_passages[2].title}
Text: {retrieved_passages[2].text}

Title: {retrieved_passages[3].title}
Text: {retrieved_passages[3].text}

Title: {retrieved_passages[4].title}
Text: {retrieved_passages[4].text}

Input: {claim} True or False?
Output:
    
    \end{lstlisting}
    
    \caption{Claim verification prompt adapted from \citet{min-etal-2023-factscore}, used to
classify each atomic claim as supported or unsupported given retrieved Wikipedia passages.}
    \label{fig:claim_verification_prompt}
    
\end{figure}

\begin{figure*}[!t]
    \centering
    \begin{minipage}{0.9\textwidth}

        \begin{tabular}{|l|p{10cm}|}
            \hline
            \textbf{Example prompt 1}
                & \begin{lstlisting}[basicstyle=\scriptsize\ttfamily,breaklines=true]
<USER PROMPT>
Write a short biography of Lokesh Kanagaraj.
<ASSISTANT RESPONSE>
Lokesh Kanagaraj (born 14 March 1986) is an Indian film director, screenwriter, and producer known for his work in Tamil cinema. He began his career with the short film in the 2016 anthology Aviyal and directed his first feature film Maanagaram (2017). He created the Lokesh Cinematic Universe (LCU) with the success of Kaithi (2019).

He directed Master (2021), a standalone film featuring Vijay, followed by Vikram (2022), the second LCU installment. His film Leo (2023), the third LCU installment, became a major success.
<INSTRUCTION>
Write a two times longer response. Use all the information from the given response, and include new facts to create a more detailed response.
<LONGER ASSISTANT RESPONSE>
                    \end{lstlisting}
                \\
            \hline
            \textbf{Example response 1}
                & \begin{lstlisting}[basicstyle=\scriptsize\ttfamily,breaklines=true]
Lokesh Kanagaraj is an Indian filmmaker born on March 14, 1986. He has established himself as a prominent figure in Tamil cinema, known for his work as a director, screenwriter, and producer.

Kanagaraj's career began with the short film segment in the 2016 anthology film "Aviyal." This marked his entry into the filmmaking world. Subsequently, he directed his first feature film, "Maanagaram," released in 2017.

"Maanagaram" garnered critical acclaim and established Kanagaraj as a promising new talent in Tamil cinema. He further solidified his position with the 2019 film "Kaithi." The success of "Kaithi" led to the creation of the Lokesh Cinematic Universe (LCU), a series of interconnected films sharing a common narrative and characters.

Following "Kaithi," Kanagaraj directed "Master" (2021), a standalone film featuring the prominent Tamil actor, Vijay. While not directly part of the LCU, "Master" showcased Kanagaraj's distinct style and narrative techniques.

In 2022, Kanagaraj delivered "Vikram," the second installment in the LCU. "Vikram" received widespread critical and commercial success, further expanding the LCU's reach and solidifying Kanagaraj's position as a leading filmmaker in Tamil cinema.

The third installment in the LCU, "Leo," was released in 2023. "Leo" achieved significant commercial success, further enhancing the popularity and impact of the LCU within the Tamil film industry.

Lokesh Kanagaraj's contributions to Tamil cinema include his innovative storytelling, distinctive visual style, and his successful creation and development of the Lokesh Cinematic Universe. He continues to be a prominent figure in the industry, with his work garnering critical acclaim and a dedicated fan following.
                    \end{lstlisting}
                \\
            \hline
            \textbf{Prompt}
                & \begin{lstlisting}[basicstyle=\scriptsize\ttfamily,breaklines=true]
<USER PROMPT>
{prompt}
<ASSISTANT RESPONSE>
{original_response}
<INSTRUCTION>
Write a two times longer response. Use all the information from the given response, and include new facts to create a more detailed response.
<LONGER ASSISTANT RESPONSE>
                \end{lstlisting}
                \\
            \hline
            \end{tabular}

    \caption{Few-shot prompt for the brainstorming step in Evidence Rewrite, which instructs the base model to generate a longer, more detailed response given an initial short generation. One of two few-shot examples is shown.}
    \label{fig:brainstorm_example}

\end{minipage}
\end{figure*}

\begin{figure}[!t]

\begin{lstlisting}[
    basicstyle=\scriptsize\ttfamily,
    linewidth=.95\linewidth,
    frame=single,
    breaklines=true,
    breakindent=0pt,
    framesep=3mm,
    numbers=none,
]
[START USER PROMPT]
{prompt}
[END USER PROMPT]

The following claims are factually correct:
1. {claim[0]}
2. {claim[1]}
3. {claim[2]}
...
n. {claim[n]}

Generate a helpful response to the user prompt by summarizing the given claims without adding new information. If there are not enough information to create a helpful response, tell the user that you cannot answer because you do not know.
\end{lstlisting}

\caption{Summarization prompt used in the final rewriting step of Evidence Rewrite, which
instructs the rewriter model to compose a fluent response from the set of externally
verified claims only, without introducing new information.}
\label{fig:summarization_prompt_template}

\end{figure}
\begin{figure*}[!t]
\centering
\begin{minipage}{0.98\textwidth}
\begin{lstlisting}[
    basicstyle=\tiny\ttfamily,
    breaklines=true,
    frame=single,
    breakindent=0pt,
    framesep=1mm,
    aboveskip=0pt,
    belowskip=0pt,
]
You are a TRIVIA QUIZ MASTER.

You are preparing questions for a quiz show.
You are given a CLAIM (a piece of information).

==================================================
INPUT (VISIBLE FOR YOU ONLY)
==================================================

Original question:
<Original question>

Original response:
<Original response>

Claim:
<Claim text>

==================================================
STEP 1 — AUDIT: CAN THIS BE A GOOD TRIVIA QUESTION?
==================================================

A claim is SUITABLE only if ALL are true:
• It contains a specific fact about the real world that can be asked.
• It is generally true, without requiring any context from the original question or response.
• A participant would need to have learned this fact.
• It is clearly understandable and complete on its own, without any context from the original question or response.
• It is not opinion based or an arbitraty definition.
• It is not obvious, generic, or common-sense.
• I can ask about it without revealing the answer.

REJECT the claim if:
• It describes a vague belief, attitude, or opinion.
• It depends on additional information or original context to make sense.
• It does not carry enough factual information to be meaningful on its own.
• It would not make a meaningful quiz question.

When unsure → NOT suitable.

==================================================
STEP 2 — QUESTION GENERATION (ONLY IF SUITABLE)
==================================================

Generate EXACTLY 5 questions.
The participant will see ONLY each generated question. They will NOT see the claim.

Each question must function as a knowledge probe:
• A minimal correct answer would necessarily include ALL factual information in the claim.
• The question must NOT teach, hint, or guide toward the answer.
• The question must stand fully on its own.

Strictly adhere to these rules:
1. No answer leakage (no distinctive names, numbers, dates, or phrases from the claim unless unavoidable).
2. Do not closely paraphrase the claim.
3. Do not introduce new facts or details not in the claim.
4. If the claim has multiple factual components, each question must require ALL components.
5. Use varied, natural phrasings (avoid trivial rewrites).
6. Avoid overly constrained questions that allow guessing without knowledge.
7. Must be answerable on its own, without needing to see the original claim, question or response.

If you cannot satisfy ALL these rules, mark the claim as not suitable.

==================================================
OUTPUT FORMAT
==================================================

You must output ONLY valid JSON. 

{{
  "audit_checklist": {{
    "is_independent_of_context": "Explain if this claim is meaningful and generally true on its own, without any context from the original question or response.",
    "contains_specific_fact": "Explain if this is a hard fact or just an opinion, attitude, or arbitrary definition.",
    "is_not_obvious": "Explain if this is actual trivia or just trivial common sense.",
    "can_be_asked_for": "Explain if this claim can be turned into a meaningful trivia question."
  }},
  "suitability_reason": "Your final decision on why the claim passes or fails suitability.",
  "is_suitable": true or false,
  "questions": []
}}

If "is_suitable" is false, the "questions" list MUST be empty [].
\end{lstlisting}
\end{minipage}
\caption{Prompt template for generating probing questions from atomic claims in the Recall
Rewrite pipeline. For each claim, the model first audits whether the claim is suitable for
probing, then generates five context-independent questions designed to test whether the base
model can recall the claim without being guided toward the answer.}
\label{fig:claim_question_rewrite}
\end{figure*}

\begin{figure*}[!t]
\begin{minipage}{0.9\textwidth}
\centering
\begin{lstlisting}[
    basicstyle=\scriptsize\ttfamily,
    breaklines=true,
    frame=single,
    breakindent=0pt,
    framesep=3mm,
]
You are a redaction editor. Your task is to prepare responses for release by removing restricted information.

You will receive:
- a question
- an original response
- a list of all information from the original response
- a list of restricted information

Your task is to rewrite the original response so that the final response does not contain or reveal any of the restricted information.

Editing rules:
1. The final response must not contain, imply, or reveal any of the restricted information.
2. This includes explicit statements, paraphrases, partial references, or indirect implications of the restricted information.
3. Preserve as much of the original response as possible. Make the smallest edits necessary to remove the restricted information.
4. Do not add new information, facts, or explanations that were not already present in the remaining parts of the original response.
5. Prefer local edits (removing or slightly rewriting only the affected parts) rather than rewriting the entire response.
6. Apart from removing restricted information, keep the wording, structure, and style of the original response as unchanged as possible.

Insufficient information rule:
If removing the restricted information leaves too little information to meaningfully answer the question, do not attempt to repair or guess the answer. Instead, return:
"I cannot answer as I have insufficient information."

Output format:
Return only valid JSON in the following format:

{{
    "insufficient_information": true or false,
    "answer": "your revised answer here"
}}

If insufficient_information is true, the "answer" field must contain exactly:
"I cannot answer as I have insufficient information."

Input:

Question:
<Original question>

Original Response:
<Original Response>

All Information:
<List of all claims>

Restricted Information:
<List of all claims to remve>
\end{lstlisting}
\end{minipage}
\caption{Prompt template for the response rewriting step in the Recall Rewrite pipeline,
which removes claims classified as unknown while preserving the structure and wording of the
original response. If too little information remains after redaction, the model returns a
fixed abstention string instead.}
\label{fig:template_redaction}
\end{figure*}
\begin{figure*}[!p]
\centering
\begin{minipage}{0.98\textwidth}
\centering
\begin{lstlisting}[
    basicstyle=\scriptsize\ttfamily,
    breaklines=true,
    frame=single,
    breakindent=0pt,
    framesep=2mm,
    aboveskip=0pt,
    belowskip=0pt,
]
You are a CURRICULUM DESIGNER preparing students for an exam.

The exam consists of ONE question.
Students must produce an answer that is semantically equivalent to the provided gold answer.

You are given:
- The exam question
- The gold answer

Your task is to identify ALL knowledge that must be taught so that a student could reliably produce a correct answer equivalent to the gold answer.

Your student:

- Has basic language ability.
- Has basic everyday common-sense knowledge.
- Does NOT possess any domain-specific, cultural, procedural, or specialized knowledge.
- Does NOT possess knowledge of specific conventions unless explicitly taught.

You must determine what knowledge needs to be taught.

==================================================
DEFINITION OF KNOWLEDGE
==================================================

Knowledge can be:

1. Explicit declarative knowledge  
   - Concrete factual statements stated in the answer.

2. Background world knowledge  
   - Facts assumed but not stated.

3. Procedural knowledge  
   - Methods, techniques, patterns, or processes required to construct the answer.

4. Cultural or genre knowledge  
   - Social norms and traditions, stylistic conventions, or domain-specific practices relied upon.

5. Structural knowledge  
   - Required format, logical organization, or compositional constraints necessary to produce the answer.

6. Relational knowledge  
   - Necessary relationships between facts or concepts required for correctness.

==================================================
NECESSITY TEST
==================================================

Include a knowledge item ONLY if:

- If the student lacked this knowledge, they would likely fail to produce a correct answer.
- The knowledge is specific and learnable.
- The knowledge goes beyond basic language competence and everyday common sense.
\end{lstlisting}
\end{minipage}
\caption{Prompt template for knowledge decomposition in the Recall Rewrite pipeline
(part 1 of 2).}
\label{fig:confidence_rewrite_decomp}
\end{figure*}

\begin{figure*}[!p]
\ContinuedFloat
\iflatexml\addtocounter{figure}{-1}\fi%
\centering
\begin{minipage}{0.98\textwidth}
\centering
\begin{lstlisting}[
    basicstyle=\scriptsize\ttfamily,
    breaklines=true,
    frame=single,
    breakindent=0pt,
    framesep=2mm,
    aboveskip=0pt,
    belowskip=0pt,
]

Do NOT include:

- Basic ability to understand language.
- Basic reasoning ability.
- Extremely general common knowledge (e.g., "movies are watched on screens").
- Trivial restatements of the question.

==================================================
OUTPUT REQUIREMENTS
==================================================

1. Output ONLY a JSON list of knowledge items.

2. Each knowledge item must:
   - Be a complete declarative sentence.
   - Express exactly ONE atomic fact.
   - Be generally true.
   - Be understandable without the original question or answer.
   - Contain a clear subject and predicate.
   - Avoid pronouns whose referent depends on other items.

3. Do NOT:
   - Use meta-language such as "The student must know" or "This answer requires".
   - Describe abilities or vague knowledge areas (e.g., "Knowledge of Flask").
   - Combine multiple independent facts into one sentence.
   - Refer to the question, answer, exam, or student.

4. Knowledge items must be atomic. 
   - If it contains multiple facts, split it into separate items.
   - If it can be split into two items without losing its meaning, its not atomic.

==================================================
INPUT
==================================================

Question:
<Question>

Gold Answer:
<Answer>

==================================================
OUTPUT FORMAT
==================================================

[
  "Knowledge item 1",
  "Knowledge item 2",
  ...
]
\end{lstlisting}
\end{minipage}
\caption{Prompt template for knowledge decomposition in the Recall Rewrite pipeline, which
extracts all atomic knowledge items a model would need to produce a given gold response,
covering explicit facts, background knowledge, procedural, cultural, structural, and
relational knowledge. Only items whose absence would cause the model to fail are retained
(continued).}
\end{figure*}

\begin{figure*}[!t]
\begin{minipage}{0.9\textwidth}
\centering
\begin{lstlisting}[
    basicstyle=\scriptsize\ttfamily,
    breaklines=true,
    frame=single,
    breakindent=0pt,
    framesep=3mm,
]
You are a strict quiz grader.

A student submitted an ANSWER.
There is a REQUIRED FACT that the answer is supposed to contain.

Your job is to decide whether the ANSWER successfully includes the REQUIRED FACT.

========================
HOW TO DECIDE
========================

1. Mentally break the REQUIRED FACT into its smallest factual elements.

2. Check whether each element appears explicitly in the ANSWER.

Use these rules:

ENTAILS:
- Every factual element of the REQUIRED FACT is clearly and explicitly stated in the ANSWER.
- Paraphrasing is allowed.
- The ANSWER is at least as specific as the REQUIRED FACT.

CONTRADICTS:
- The ANSWER explicitly conflicts with any factual element of the REQUIRED FACT.
- A different number, date, entity, or opposite claim counts as a contradiction.

UNRELATED:
- Any factual element is missing.
- The ANSWER is incomplete or too vague to recover the full fact.
- The ANSWER discusses the topic but does not clearly state the required fact.
- The ANSWER neither clearly supports nor contradicts the fact.

Important:
- The ANSWER may contain additional information.
- Extra details are allowed and should NOT affect the label.
- Only mark CONTRADICTS if the additional information directly conflicts with the REQUIRED FACT.
- Do not use outside knowledge. Only use what is written.
- If any required element is missing, the label cannot be ENTAILS.

========================
OUTPUT
========================

Respond with exactly one of:

ENTAILS
CONTRADICTS
UNRELATED

========================
TASK
========================

REQUIRED FACT:
"<Claim Text>"

ANSWER:
"<Model Answer>"
\end{lstlisting}
\end{minipage}
\caption{Prompt template for the entailment check in the Recall Rewrite pipeline, which
classifies each base model response to a probing question as \texttt{ENTAILS},
\texttt{CONTRADICTS}, or \texttt{UNRELATED} with respect to the original claim, based
solely on what is explicitly stated in the response.}
\label{fig:template_entailment_check}
\end{figure*}
\begin{figure*}[!t]
\begin{minipage}{0.9\textwidth}
\centering
\begin{lstlisting}[
    basicstyle=\scriptsize\ttfamily,
    breaklines=true,
    frame=single,
    breakindent=0pt,
    framesep=3mm,
]
<|im_start|>system
You are a question answering system.
Always provide a clear and complete answer to the user's question.
Do not repeat or restate the question.
Each answer should be a full sentence that fully addresses the question.<|im_end|>

<|im_start|>user
What form of government is characterized by power being concentrated in the hands of a small group of people?<|im_end|>
<|im_start|>assistant
An oligarchy is a form of government in which political power is held by a small, elite group.<|endoftext|>

<|im_start|>user
If the current time is 15:00 UTC, what would the local time be in a region that operates at UTC+2?<|im_end|>
<|im_start|>assistant
The local time would be 17:00, because the region is two hours ahead of Coordinated Universal Time.<|endoftext|>

<|im_start|>user
In economics, what term describes a market situation where the quantity supplied exceeds the quantity demanded?<|im_end|>
<|im_start|>assistant
This situation is known as a surplus, where more goods are produced than consumers are willing to buy at a given price.<|endoftext|>

<|im_start|>user
Which biological process enables cells to convert glucose into usable energy?<|im_end|>
<|im_start|>assistant
Cells use cellular respiration, a metabolic process that converts glucose into usable energy in the form of ATP.<|endoftext|>

<|im_start|>user
Many administrative systems assign fixed-length numerical identifiers to individuals. What property ensures that each identifier refers to only one person?<|im_end|>
<|im_start|>assistant
This property is uniqueness, meaning each identifier is assigned to a single individual and cannot be shared with others.<|endoftext|>

<|im_start|>user
{Question}<|im_end|>
<|im_start|>assistant
\end{lstlisting}
\end{minipage}
\caption{Few-shot prompt used to query the base model with probing questions in the Recall
Rewrite pipeline. The five examples establish a question-answering format and encourage
complete, self-contained answers without restating the question.}
\label{fig:inference}
\end{figure*}

\end{document}